\documentclass[letterpaper,11pt]{article}

\usepackage[margin=1in]{geometry}
\usepackage{graphicx}
\usepackage[hyphens]{url}
\usepackage{natbib}
\usepackage{caption}
\usepackage{booktabs}
\usepackage{array}
\usepackage{tabularx}
\usepackage{amsmath,amssymb}
\usepackage{hyperref}
\usepackage{enumitem}
\usepackage{float}
\usepackage{placeins}
\usepackage{microtype}
\usepackage{silence}
\usepackage{xcolor}
\usepackage{tikz}
\usepackage{algorithm}
\usepackage{algorithmic}
\usepackage{textcomp}
\usetikzlibrary{positioning,arrows.meta,calc,fit,backgrounds}

\hypersetup{
  colorlinks=true,
  linkcolor=blue,
  citecolor=blue,
  urlcolor=blue,
  pdfborder={0 0 0},
  bookmarksdepth=2
}

\makeatletter
\let\oldaddcontentsline\addcontentsline
\def\addcontentsline#1#2#3{%
  \def\@tmpa{#1}\def\@tmpb{toc}%
  \ifx\@tmpa\@tmpb
    \def\@tmpa{#2}%
    \def\@tmpb{section}\ifx\@tmpa\@tmpb\oldaddcontentsline{#1}{#2}{#3}\fi
    \def\@tmpb{subsection}\ifx\@tmpa\@tmpb\oldaddcontentsline{#1}{#2}{#3}\fi
  \fi}
\makeatother

\AtBeginDocument{\setcounter{tocdepth}{2}}

\newcolumntype{L}[1]{>{\raggedright\arraybackslash}p{#1}}
\newcolumntype{Y}{>{\raggedright\arraybackslash}X}

\definecolor{cream}{HTML}{FAF5EC}
\definecolor{parchment}{HTML}{F1E8D9}
\definecolor{warmgray}{HTML}{8B8175}
\definecolor{warmgrayfill}{HTML}{E4DDCF}
\definecolor{ochre}{HTML}{A87C3E}
\definecolor{ochrefill}{HTML}{EEDCB9}
\definecolor{sage}{HTML}{7C8A66}
\definecolor{sagefill}{HTML}{DFE3CE}
\definecolor{terra}{HTML}{A2663F}
\definecolor{ink}{HTML}{38332B}

\newcommand{\yfull}{\ensuremath{\checkmark}}
\newcommand{\yno}{--}
\newcommand{\ypart}{\tikz[baseline=-0.65ex]{\draw[ink,line width=0.5pt] (0,0) circle (0.52ex); \fill[ink] (0,0) -- (90:0.52ex) arc (90:270:0.52ex) -- cycle;}}
\newcommand{\paperlink}[2]{\hyperlink{cite.#1}{#2}}

\newlength\titlebox
\newcommand{\paperauthornames}{Weicheng Ye, Youran Sun, Xingyu Ren,
  Shunyao Yu, Chugang Yi, and Haizhao Yang}
\newcommand{\makesupplementtitle}{%
  \begin{center}
  {\LARGE\bfseries Supplementary Material for\\
  AgonAlpha: Autonomous Alpha Discovery via Prompt Economy and Scalable Agentic Search\par}
  \vskip 0.25in
  {\large\bfseries \paperauthornames\par}
  \end{center}
  \vskip 0.35in
}

\title{\Large\bfseries AgonAlpha: Autonomous Alpha Discovery via Prompt Economy and Scalable Agentic Search}

\author{%
Weicheng Ye\textsuperscript{1,*,\textdaggerdbl},
Youran Sun\textsuperscript{2,*},
Xingyu Ren\textsuperscript{1,*},\\[0.25em]
Shunyao Yu\textsuperscript{1},
Chugang Yi\textsuperscript{2},
Haizhao Yang\textsuperscript{2,3,\textdagger}\\[0.75em]
\textsuperscript{1}Department of Physics, The Chinese University of Hong Kong, Hong Kong SAR, China\\
\textsuperscript{2}Department of Mathematics, University of Maryland, College Park, MD, USA\\
\textsuperscript{3}Department of Computer Science, University of Maryland, College Park, MD, USA
}
\date{}

\begin{document}

\begingroup
\renewcommand{\thefootnote}{\fnsymbol{footnote}}
\maketitle
\footnotetext[1]{Equal contribution.}
\footnotetext[2]{\texttt{hzyang@umd.edu}.}
\footnotetext[3]{\texttt{victoryeofphysics@gmail.com}.}
\endgroup

\tableofcontents
\vspace{1em}

\begin{abstract}
Language models can propose many plausible trading factors, but an autonomous research system must also allocate its evaluation budget, verify its own evidence, and preserve how each candidate was produced.
We present AgonAlpha, an architecture that searches over frozen research artifacts---hypotheses, executable expressions, platform evidence, rationales, and review status---rather than formulas alone.
To our knowledge, AgonAlpha is the first alpha-mining system to combine verified artifact search, a fresh-context adversarial reviewer with re-execution and veto authority, and pending-aware parallel budget allocation, together with a complete public evidence trail.
Independent deployments on WorldQuant BRAIN produced SPECTACULAR-grade alphas across five users and six model backends, with Fitness reaching 9.50 and Sharpe reaching 3.48, while retaining prompt-to-expression provenance for every submission.
\end{abstract}
\section{Introduction}

With the increasing adoption of agentic workflows, autonomous alpha discovery is not merely about generating plausible formulas by large language models (LLMs), but also designing a comprehensive system of evaluation and self-improvement. 
The system should go beyond systematically and consistently generating plausible alphas: it should decide which families of alpha ideas deserve the limited number of expensive platform evaluation opportunities, ensure that the recorded evidence corresponds to the evaluated candidates, and maintain sufficient state to enable reliable reconstruction of results after each run. 
AlphaBench shows that language models still struggle with zero-shot factor ranking, even though they are relatively effective at generating executable factors \cite{luo2026alphabench}. Meanwhile, the performance of published anomalies has also weakened substantially over time. In the post-2005 non-micro-cap sample analyzed by Chen and Welch, the median return is only around 7 basis points per month—an effect size small enough to be explained away by statistical luck or modest transaction costs \cite{chen2026decay}. Moreover, a systematic review of 30 LLM-based trading papers reveals a recurring gap in research practice: while model architectures are often described in detail, crucial aspects such as point-in-time data controls, train-test splits, out-of-sample evaluation, transaction costs, turnover assumptions, execution details, and artifact availability are frequently underreported \cite{yao2026beyond}.
Together, these findings locate the central challenge at the level of the automated alpha-discovery loop: preserving evidence, allocating costly evaluations across competing ideas, and separating proposal from verification.

\begin{figure*}[t]
\centering
\includegraphics[width=\textwidth]{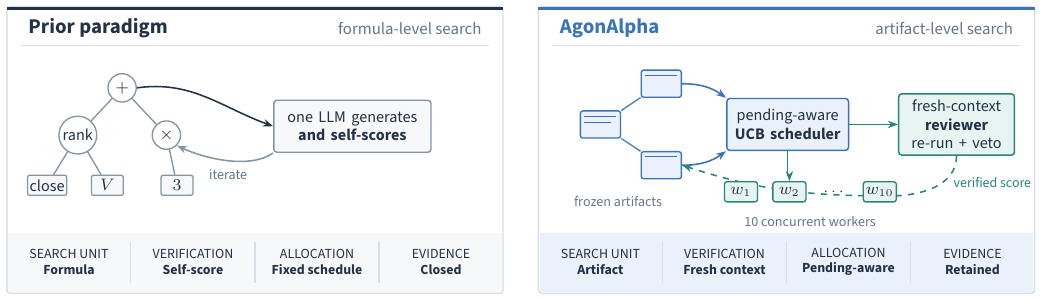}
\caption{Two paradigms for LLM-driven alpha mining.
\emph{Left}: formula-level search with self-scoring and a closed trail.
\emph{Right}: AgonAlpha searches over research artifacts subject to fresh-context review with a pending-aware scheduler validated under 10 concurrent workers.}
\label{fig:paradigm}
\end{figure*}

For this purpose, we apply the \emph{artifact-level} discovery; compared with previous works, many of which are at the \emph{formula-level}, there are four distinctive architectural choices, as shown in Figure~\ref{fig:paradigm}.

\begin{itemize}
\item \textbf{Search unit.}
Existing systems typically treat a formula as the basic unit of discovery. However, a formula alone does not preserve the hypothesis that motivated it, the alternative directions that were explored and rejected, or the objections that were addressed along the way. We instead treat discovery as a lineage of research decisions, preserving the context needed to understand why a candidate was generated and how it evolved.

\item \textbf{Verification mechanism.}
Existing systems often rely on self-evaluation procedures or scalar performance thresholds. While these mechanisms are effective for ranking candidates, they do not independently verify whether the supporting evidence is consistent with the evaluated candidate. We introduce explicit verification procedures to audit the correspondence between generated candidates, evaluation records, and reported results.

\item \textbf{Resource allocation.}
Existing systems commonly allocate search budgets according to fixed schedules. However, when external platform evaluations require substantial time and multiple research directions proceed concurrently, static allocation cannot adaptively redirect resources toward promising research lineages or account for evaluations already in progress. We instead treat evaluation budget allocation as an active decision-making problem, allowing the system to prioritize promising directions while coordinating ongoing experiments.

\item \textbf{Evidence preservation.}
Existing systems often provide incomplete evidence trails. The audit of 30 LLM-based trading papers shows that critical execution and evaluation details needed to interpret reported results are frequently less accessible than descriptions of the agent architecture itself \cite{yao2026beyond}. We therefore emphasize the preservation of complete research artifacts and execution history, enabling results to be reproduced and independently assessed.

\end{itemize}

Together, these differences highlight a broader distinction: these are not merely limitations of individual factor-generation methods, but missing interfaces required for building a fully autonomous discovery system.

AgonAlpha implements these interfaces through three coordinated components: a proposer, a fresh-context reviewer, and a pending-aware scheduler. The proposer commits each candidate as an evidence-bearing artifact; the reviewer independently checks the artifact, may rerun the associated platform evaluation, and can reject unsupported or fabricated claims; and the scheduler allocates each subsequent invocation to a research lineage while accounting for evaluations already in progress. This architecture follows the \emph{Agon philosophy} and the Prompt Economy principle \cite{sun2026agon,sun2026perspectivegap}. Five co-authors deployed AgonAlpha on WorldQuant BRAIN using multiple model backends, producing 60 submissions, of which 17 received SPECTACULAR grades; the best observed Fitness and Sharpe ratio were 9.50 and 3.48, respectively. We present detailed analyses of five representative cases and release the prompts, decisions, reviews, platform evidence, and factor expressions for all 60 submissions, with the complete collection provided in the supplementary material.

On top of the design of the system, the design of the evaluation is itself a key systems consideration. Existing approaches, including AlphaForge, AlphaJungle, and FactorMiner, evaluate generated factors through author-controlled academic pipelines, even when they introduce temporal or cross-market holdouts \cite{shi2025alphaforge,shi2026alphajungle,wang2026factorminer}. Such protocols improve the separation between discovery and evaluation data, but the evaluator remains part of the system being designed. Our approach instead adopts WorldQuant BRAIN as an external evaluation environment: a platform that independently determines the data, simulation rules, metrics, submission constraints, and final scores, with an incentive mechanism for qualifying research consultants \cite{WQ}. Therefore, our results measure not only factor discovery capability but also the ability of an autonomous research system to operate under a fixed, externally governed evaluation regime.

\noindent\textbf{Contributions.}
\begin{itemize}
\item \textbf{C1 --- An integrated architecture for auditable alpha discovery.}
To our knowledge, AgonAlpha is the first alpha-mining system to combine verified artifact search, adversarial review authority, pending-aware parallel budget allocation, and a complete public evidence trail.
Its compact realization applies the six Agon principles---Prompt Economy, Minimal Prompts, Future-Facing, Zero-Code, OmniDisciplinary, Massive Parallelism---through two roles and 101 lines of role prompt \cite{sun2026agon,sun2026perspectivegap}.
\item \textbf{C2 --- Trajectory-level search over evidence-bearing artifacts.}
The search unit is a frozen research artifact containing a hypothesis, executable expression, platform evidence, economic rationale, and review status.
The scheduler therefore allocates work across research lineages rather than isolated formula edits.
\item \textbf{C3 --- Adversarial review with re-execution and veto authority.}
The reviewer is separately routed and invoked in a fresh context, may re-run platform evidence, and can set the scheduler reward to zero when it verifies fabrication.
Other risk findings remain attached to the artifact as inspectable warnings.
\item \textbf{C4 --- Pending-aware parallel budget allocation.}
Across lineages, a pending-aware MCTS scheduler combines progressive widening, percentile rewards, backpressure, and a root fallback so that in-flight work affects subsequent allocation.
Within each pipeline, a halving tournament concentrates simulations on surviving candidates (\S\ref{sec:allocation}).
\item \textbf{C5 --- Trace-complete multi-user deployment validation.}
Five users and multiple model backends produced 60 externally graded submissions, including 17 SPECTACULAR alphas and Fitness reaching 9.50.
We release every prompt, search decision, platform record, review, and executable expression.
\end{itemize}
\section{Related Work}

\subsection{Axis I: The Unit of Search}
Most alpha-mining agents search over formulas.
AlphaJungle applies Monte Carlo Tree Search (MCTS) to formula expression trees, using one LLM to generate and self-score candidates \cite{shi2026alphajungle}.
It asks \emph{where to add the next operator}; we ask \emph{which lineage deserves the next budget}.
QuantaAlpha evolves trajectories as genetic material under a fixed five-round schedule; our artifacts are evidence units---frozen, adversarially gated, UCB-budgeted \cite{quantaalpha2026}.
FAMA chains experiences linearly and reports hallucinations in its own evaluation \cite{li2024fama}, directly motivating C3.

\subsection{Axis II: The Verification Regime}
FactorMiner \cite{wang2026factorminer}, the nearest competitor, has the same role count but the opposite organizing principle.
Its roles divide labor functionally (generator plus deterministic tool evaluation); ours divide adversarially (proposer plus attacker).
Its Ralph Loop is a linear cycle with no tree, no UCB, and no cross-lineage allocation.
Its validation is a \emph{tool}---deterministic thresholds on IC and correlation---while ours is an \emph{authority} with re-run and veto rights.
FactorMiner publishes its factor library but keeps code and prompts closed.
FactorMAD's debate is cooperative with no veto, and admission returns to a deterministic cutoff \cite{factormad2025}.
Agora's panel disperses power, so no single verifier holds both re-run and veto rights.
Its own analysis concludes that adding agents merely relocates rather than resolves the failure of self-confirmation \cite{agora2026}.
CogAlpha fields 21 agents at roughly 500 H100-hours per run with same-model quality control and closed prompts \cite{cogalpha2025}.
Beyond Prompting's fixed threshold gate fails in public \cite{huang2026beyond}, and AlphaCrafter documents the backtest-to-live cliff \cite{alphacrafter2026,atlas2025}.
AlphaBench is not a competing system.
Its finding that LLM factor evaluation is near-random provides the strongest published motivation for C3 \cite{luo2026alphabench}.

\subsection{Axis III: Budget Allocation}
R\&D-Agent-Quant alternates factor and model development under a two-arm bandit---arms, not tree nodes, with no notion of in-flight work \cite{li2025rdagent}.
AlphaGen reports exploration stagnation, while QF-REINFORCE's algorithm-layer patch leaves its CSI500 RankIC below AlphaGen's without explaining the gap \cite{yu2023alphagen,zhao2025qfr}.
AlphaForge instead runs a hardcoded pipeline over a frozen factor pool \cite{shi2025alphaforge}.
QuantEvolver addresses the same problem by encoding experience in model weights through GRPO \cite{zhang2026quantevolver}.
AgonAlpha instead searches at inference time, which is the only option available to platform users without weight access.
Table~\ref{tab:comparison} maps the field onto the four failure axes of \S1.
The F4 column shows the domain norm: no system releases prompts, search decisions, review text, and exact executable expressions together---most expose code only.
C5 addresses this gap with a release that covers all five dimensions of the 30-paper audit (\S\ref{sec:evaluation}).

\nocite{weng2026alphalogics}
\begin{table*}[!t]
\centering
\scriptsize
\setlength{\tabcolsep}{4pt}
\begin{tabularx}{\textwidth}{@{}L{3.1cm}YYYY@{}}
\toprule
System & Search unit (F1) & Verification (F2) & Allocation (F3) & Open trail (F4) \\
\midrule
\paperlink{shi2025alphaforge}{AlphaForge (AAAI'25)} & formula & \ypart~IC/correlation thresholds & \ypart~hardcoded pipeline & \ypart~code only \\ \hline
\paperlink{yu2023alphagen}{AlphaGen (KDD'23)} & formula & \ypart~RL reward & \ypart~RL policy & \ypart~code only \\ \hline
\paperlink{zhao2025qfr}{QF-REINFORCE} & formula & \ypart~IR reward shaping & \ypart~RL policy & \yno \\ \hline
\paperlink{li2024fama}{FAMA (ACL'24)} & formula & \yno~RankIC threshold & \yno~linear chain & \yno \\ \hline
\paperlink{factormad2025}{FactorMAD (ICAIF'25)} & formula & \ypart~cooperative debate $+$ cutoff & \yno~fixed rounds & \yno \\ \hline
\paperlink{shi2026alphajungle}{AlphaJungle (AAAI'26)} & formula edits & \ypart~LLM self-score & \ypart~serial UCT & \yno \\ \hline
\paperlink{li2025rdagent}{R\&D-Agent-Q (NeurIPS'25)} & factor $+$ model & \ypart~thresholds $+$ backtest & \ypart~two-arm bandit & \ypart~code $+$ prompts \\ \hline
\paperlink{wang2026factorminer}{FactorMiner (ICLR'26)} & \ypart~formula $+$ traj.\ log & \ypart~deterministic pipeline & \ypart~memory-biased sampling & \ypart~factor library \\ \hline
\paperlink{quantaalpha2026}{QuantaAlpha} & \ypart~trajectory (genetic) & \ypart~self-consistency & \yno~fixed schedule & \ypart~code only \\ \hline
\paperlink{cogalpha2025}{CogAlpha} & code formulas & \ypart~same-model QC chain & \ypart~$(\mu,\lambda)$ evolution & \yno \\ \hline
\paperlink{agora2026}{Agora} & formula $+$ scorers & \ypart~panel (dispersed power) & \yno~serial rounds & \ypart~harness/prompts closed \\ \hline
\paperlink{huang2026beyond}{Beyond Prompting} & formula (grammar) & \yno~fixed threshold gate & \yno~linear loop & \yno \\ \hline
\paperlink{zhang2026quantevolver}{QuantEvolver} & formula (DSL) & \ypart~backtest reward & \ypart~GRPO training & \ypart \\ \hline
\paperlink{weng2026alphalogics}{AlphaLogics} & logic $+$ formula & \ypart~aggregate statistics & \ypart~dual loop & \ypart \\ \hline
\paperlink{alphacrafter2026}{AlphaCrafter} & factor library & \ypart~regime/risk gates & \yno~pipeline & \ypart \\ \hline
\midrule
\textbf{AgonAlpha (ours)} & \textbf{\yfull~verified artifact} & \textbf{\yfull~re-run $+$ veto, one agent} & \textbf{\yfull~pending-aware UCB} & \textbf{\yfull~full public trail} \\ \hline
\bottomrule
\end{tabularx}
\caption{Representative formulaic alpha-mining systems on the four failure axes of \S1 (F1 search unit, F2 verification regime, F3 budget allocation, F4 open evidence trail).
\yfull{} = present, \ypart{} = partial/indirect, \yno{} = absent or unconfirmed.
Performance numbers are excluded: universes, delays, and evaluation windows differ across studies.}
\label{tab:comparison}
\end{table*}
\section{Minimal Instance of the Agon Philosophy}
\label{sec:philosophy}

The Agon philosophy \cite{sun2026agon} defines six design principles for autonomous research systems and grounds them in Prompt Economy \cite{sun2026perspectivegap}.
Under this view, a multi-agent system is a reusable prompt surface whose invocations must repay their cost.
AgonAlpha is the first system to instantiate all six outside of the original Agon factory.

\textbf{Prompt Economy.}
Every agent invocation costs tokens and coordination.
Systems should maximize artifact value per prompt line and per handoff.
AgonAlpha reuses two role prompts across every node of the search tree.
The halving tournament reserves later, more expensive simulations for survivors.
The tournament schedule in \S\ref{sec:allocation} measures this directly.

\textbf{Minimal Prompts.}
Shorter, fewer prompts reduce maintenance surface and coupling to specific models.
AgonAlpha's entire discovery workflow runs on 101 physical lines of role prompt: 57 for the proposer, 44 for the reviewer, plus a short dispatcher.
The full prompt set is released.

\textbf{Future-Facing.}
Prompts should describe stable roles and procedures, not patch current model weaknesses, so systems benefit from model upgrades without prompt changes.
AgonAlpha's role prompts name artifacts, checks, and stopping rules.
No prompt mentions a specific model.

\textbf{Zero-Code.}
Research logic lives in prompts, not in hardcoded pipelines.
AgonAlpha's deterministic code handles only platform I/O and tree bookkeeping.
No human wrote a factor expression or factor-specific program.

\textbf{OmniDisciplinary.}
Core agents remain domain-agnostic.
Domain knowledge enters at runtime through manuals, readings, and schemas.
AgonAlpha's role prompts contain no market-specific tokens: no ``stock,'' no ``equity,'' no ``option.''

\textbf{Massive Parallelism.}
The same prompt set should instantiate concurrently across independent research threads, limited only by platform quotas.
AgonAlpha's proposer--reviewer pipelines occupy independent worker slots; only scheduler selection and update are serialized.
The pending-aware UCB in \S\ref{sec:allocation} accounts for this concurrency.
A validated 10-worker deployment demonstrates concurrent operation.
Platform account quotas, rather than the architecture, set the ceiling.

These six principles operate through a producer--critic adversarial loop.
The proposer freezes the artifact, a separately routed fresh-context reviewer attacks it, and only then does the verified score enter the search tree (\S\ref{sec:audit}).

Two roles form the fixed point.
A single role must grade its own work, relying on the self-evaluation that AlphaBench has already shown to fail.
Adding more than two roles redistributes adversarial pressure rather than increasing it, as Agora's own analysis concluded.
Two opposed roles are the smallest structure in which an adversary can exist.
Sections~\ref{sec:system}--\ref{sec:evaluation} demonstrate that this minimal instance produces SPECTACULAR-grade alphas on WorldQuant BRAIN.
\section{The AgonAlpha System}
\label{sec:system}

\subsection{Problem Formalization}
AgonAlpha searches for cross-sectional equity alphas expressed in WorldQuant BRAIN's expression language (FASTEXPR), a domain-specific language for formulaic alpha construction.
The unit of search is an \emph{artifact} $A = (h, f, E, r, v)$: a hypothesis $h$, an executable expression $f$, evidence records $E$ (simulation inputs, metrics, annual results, submission state), an economic rationale $r$, and a verdict $v$ from the adversarial review.
A \emph{lineage} is a chain of artifacts linked by ancestry; the single search operator is $\mathrm{extend}(\ell)$, which invokes one proposer--reviewer pipeline to append a new artifact to lineage $\ell$.
Node scores enter the tree only after verification.
The scheduler's problem is budget allocation: given a budget $B$ of platform simulations and a utility $u(\cdot)$ over verified artifacts, choose extensions maximizing $\mathbb{E}\big[\sum_i u(A_i)\big]$ subject to $\sum_i \mathrm{cost}(A_i) \le B$.
The scheduler never generates expressions, runs simulations, or judges evidence; it only decides which lineage receives the next pipeline invocation.

\subsection{The Two-Role Contract}
\label{sec:roles}
The \emph{proposer} receives ancestor reports, a work directory, two sampled readings, and the platform interface.
It writes a hypothesis, generates 16 candidate formulas, and runs a halving tournament (\S\ref{sec:allocation}).
Every candidate is simulated, and near-duplicates above the self-correlation gate are removed.
Survivors are ranked by absolute platform composite score $|\mathrm{Score}|$; each pass eliminates the bottom half.
A negative-scoring finalist is sign-reflected rather than re-simulated.
For dollar-neutral cross-sectional alphas, negation reverses the signs of returns and Sharpe while leaving turnover unchanged, so the flipped candidate's metrics are derived exactly rather than re-simulated.
A monotone target constraint requires the submitted formula to beat the best ancestor score.
After the final pass, the proposer considers every eligible simulation, including eliminated candidates.
It submits the winner and freezes the full search as an alpha report.
The \emph{reviewer} receives only the work directory and platform documentation.
It runs in a fresh context on a different model route and has a single task: find grounds for rejection.
It may re-run any simulation, and verified fabrication sets the score to zero.
The two role prompts total 101 physical lines (57 proposer, 44 reviewer) in the current release.
A short dispatcher prompt connects them to the scheduler.
Reference manuals and platform documentation are runtime reading materials, not role prompts.
The full prompt history is part of the released trail.
The proposer and reviewer are routed through separate API endpoints with independent contexts; deployments used various backend combinations (see supplement for per-user configurations).

\subsection{Adversarial Verification Protocol}
\label{sec:audit}
The reviewer is not another score threshold: it must audit the evidence even when a candidate already clears every passive screen.
Table~\ref{tab:audit} states the five audit dimensions and their recorded actions.
The authority boundary is explicit: \emph{only} verified fabrication---mismatched expressions, leaked or look-ahead inputs, invented records---changes the scheduler's score, to zero.
Sign, constant, regime, and redundancy findings remain advisory records attached to the artifact.
Rules for blocking submission based on combinations of warnings are deferred to a stricter release.
The released prompt history also records one boundary migration.
Self-correlation began as a reviewer warning and moved into the proposer as a hard pre-ranking gate, currently at 0.85.
Near-duplicates are removed before ranking rather than after review.

The division of labor directly addresses the AlphaBench result.
The reviewer never estimates alpha quality---quality is scored by the platform.
The reviewer audits \emph{evidence integrity}: re-execution, record consistency, and look-ahead.
This is a consistency-checking task, not the quality-prediction task at which AlphaBench found LLM agents near-random \cite{luo2026alphabench}.
The division-of-labor argument extends only this far.
The sign and constant dimensions are judgmental; their reliability is evaluated through the reviewer audit record in \S\ref{sec:evaluation}.

\begin{table}[t]
\centering
\scriptsize
\setlength{\tabcolsep}{3pt}
\begin{tabularx}{\textwidth}{@{}L{0.18\textwidth}L{0.34\textwidth}Y@{}}
\toprule
\raggedright Dimension & \raggedright Audit question & \raggedright Recorded action \tabularnewline
\midrule
\raggedright Evidence integrity & \raggedright do expression, settings, metrics, annual table, submission state match platform records; any look-ahead? & \raggedright verified fabrication, mismatch, leakage, or look-ahead sets the score to zero \tabularnewline
\raggedright Sign logic & \raggedright does every term's direction match the stated market mechanism? & \raggedright unexplained direction $\Rightarrow$ explicit sign warning \tabularnewline
\raggedright Constant rationale & \raggedright does each unusual coefficient or window have an economic, calendar, or search-risk explanation? & \raggedright unsupported value $\Rightarrow$ explicit constant warning \tabularnewline
\raggedright Temporal stability & \raggedright are annual scores reported; does the best-to-worst annual ratio exceed five? & \raggedright missing years or ratio $>5$ $\Rightarrow$ dated regime warning \tabularnewline
\raggedright Selection risk & \raggedright did repeated search produce a near-copy of a submitted alpha, or did tuning concentrate on a narrow family? & \raggedright in-pipeline near-duplicate $\Rightarrow$ proposer-side hard gate at 0.85\newline redundancy vs.\ submitted ancestors $\Rightarrow$ advisory record \tabularnewline
\bottomrule
\end{tabularx}
\caption{The five-dimensional audit of a frozen alpha report.
Evidence failures can zero the scheduler's score; risk warnings remain visible but advisory; the correlation gate rejects near-duplicates in search.}
\label{tab:audit}
\end{table}

\begin{figure}[H]
\centering
\includegraphics[width=\textwidth]{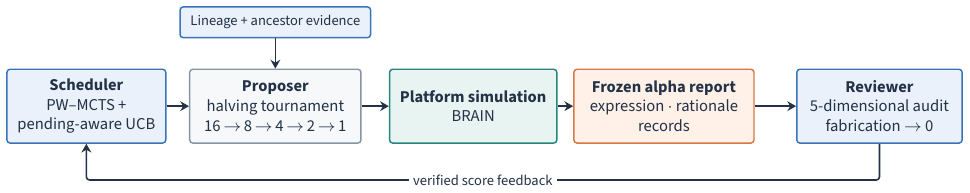}
\caption{AgonAlpha separates trajectory-level search (scheduler), candidate production (proposer), BRAIN (platform), and independent verification (reviewer), which audits the frozen report before its score is recorded.}
\label{fig:system}
\end{figure}

\subsection{Two-Level Budget Allocation}
\label{sec:allocation}
\begin{figure}[H]
\centering
\includegraphics[width=\textwidth]{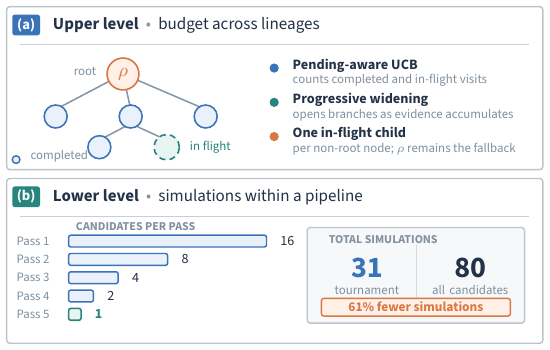}
\caption{Two-level budget allocation.
\emph{(a)} A pending-aware PW-MCTS selects which lineage receives the next pipeline; dashed nodes are in-flight.
\emph{(b)} A halving tournament concentrates 31 simulations on survivors versus the 80-simulation all-candidates schedule.}
\label{fig:budget}
\end{figure}

Evaluation in this domain is expensive and asynchronous: each simulation is billed by the platform and takes minutes, and several pipelines may be in flight at once.
AgonAlpha therefore allocates budget at two levels (Algorithm~\ref{alg:allocation}, Figure~\ref{fig:budget}).

\textbf{Upper level: pending-aware PW-MCTS across lineages.}
Write $v(n)$ and $\pi(n)$ for a node's completed and in-flight (pending) counts.
When a pipeline starts, it is credited to $\pi$ for every node on its ancestor chain and cleared when it completes.
Thus, $\pi(n)$ tracks all in-flight work beneath $n$.
A node is eligible for expansion when its completed-child count satisfies progressive widening \cite{chaslot2008pw},
\begin{equation}
\mathrm{done\_children}(n) \;<\; k\cdot\big(v(n)+\pi(n)\big)^{\alpha},
\label{eq:pw}
\end{equation}
where $k{=}1.0$ and $\alpha{=}0.5$, so the branching factor grows only as a lineage accumulates evidence.
An eligible node must also have no in-flight direct child.
One design axiom exempts the root $\rho$ from this second condition, so $\rho$ always satisfies the backpressure condition.
Equation~\eqref{eq:pw} still governs its routine expansion.
Selection descends from $\rho$ to the completed child maximizing a pending-aware upper-confidence bound \cite{kocsis2006uct} that counts in-flight candidates in both parent and child visits:
\begin{equation}
\mathrm{UCB}(c) \;=\; \frac{Q_{\mathrm{sub}}(c)}{v(c)} \;+\; C\sqrt{\frac{\ln\big(v(p)+\pi(p)\big)}{v(c)+\pi(c)}}\,,
\label{eq:ucb}
\end{equation}
\looseness=1
where $Q_{\mathrm{sub}}(c)$ is the sum of rewards over all completed artifacts in the subtree rooted at $c$, and $v(c)$ is that node's completed visit count.
The exploitation term values entire lineages using only completed evidence.
Pending counts in the exploration term prevent over-allocation to branches whose workers are still busy.
The descent stops at the first eligible node, which is expanded by creating one in-flight child.
Re-expanding a completed node forms siblings, so progressive widening bounds per-node fan-out while depth remains UCB-driven.
Under a single worker, the root becomes eligible again at $v = 2, 5, 10, 17, \ldots$.
When the UCB descent reaches a busy dead end, the fallback unconditionally expands the root $\rho$, preventing pipeline starvation.
This fallback also handles cold start: at $t{=}0$, Eq.~\eqref{eq:pw} is false for the root, so the first expansion always uses the fallback.
The eligibility rule thus gives each non-root node at most one in-flight child (backpressure).
$C{=}10.0$ matches the $[0,10]$ reward domain; all three constants appear verbatim in the released configuration.
Write \(S(A)=\mathrm{Score}(A)\).
Raw scores are non-stationary, so rewards are population percentiles via mid-rank,
\begin{equation}
r(A) \;=\;
\begin{cases}
0, & \text{fabrication verified},\\[2pt]
10\cdot\mathrm{midrank}(S(A))/N, & \text{otherwise},
\end{cases}
\label{eq:reward}
\end{equation}
with $\mathrm{midrank}=(\#\mathrm{below}+\#\mathrm{at\_or\_below})/2$ over the $N$ completed artifacts at assignment time.
A fabrication-zeroed artifact receives reward 0 regardless of its raw Score and does not enter the percentile population $N$.
Each artifact's percentile reward is computed at completion and frozen; historical rewards are not recomputed as the population grows, keeping backed-up values stable.
The percentile reward is a distribution-adaptive choice; complete scheduler logs are released for analysis.

Our selection rule extends the parallel MCTS lineage \cite{coulom2006mcts,kocsis2006uct,liu2020wuuct} with deterministic pending-count penalties rather than virtual-loss correction \cite{chaslot2008parallel,segal2010}, and composes five mechanisms for asynchronous evaluation: progressive widening, percentile rewards, backpressure, root fallback, and lineage-granularity expansion.
Together these guarantee the pipeline never starves and every lineage receives unbounded visits asymptotically \cite{kocsis2006uct,chaslot2008pw,liu2020wuuct}.

\textbf{Lower level:} \emph{elimination tournaments within a pipeline.}
Each proposer starts with 16 candidates and eliminates half per pass ($16\!\to\!8\!\to\!4\!\to\!2\!\to\!1$).
This elimination tournament follows the spirit of successive halving \cite{jamieson2016sha} and Hyperband \cite{li2017hyperband}.
Because candidates are \emph{revised} between passes, the tournament refines candidates rather than resampling fixed arms.
The platform's simulator is deterministic for a fixed expression and settings, so the planned $16{+}8{+}4{+}2{+}1 = 31$ simulations measure 31 distinct candidate versions.
The counterfactual $16{\times}5 = 80$ runs all 16 starting candidates through all five revision passes.
The ratio $2.6\times$ is an allocation count, reported as such rather than as a measured claim about tokens, money, or wall-clock time.

\begin{algorithm}[H]
\caption{Two-Level Budget Allocation}
\label{alg:allocation}
\begin{algorithmic}[1]
\REQUIRE tree $T$ rooted at $\rho$; worker pool $W$; $C{=}10.0$, $k{=}1.0$, $\alpha{=}0.5$
\STATE \textbf{eligible}$(n)$: Eq.~\eqref{eq:pw} holds \AND ($n = \rho$ \OR $n$ has no in-flight direct child)
 $\triangleright$ $\rho$ always satisfies backpressure (design axiom)
\WHILE{budget remains \textbf{and} a worker is free}
  \STATE $n \leftarrow \rho$
  \WHILE{\NOT \textbf{eligible}$(n)$ \AND $n$ has a completed child}
    \STATE $n \leftarrow \arg\max_{c} \mathrm{UCB}(c)$ over completed children of $n$ (Eq.~\eqref{eq:ucb})
  \ENDWHILE
  \IF{\textbf{eligible}$(n)$}
    \STATE \textbf{expand} $n$: assign worker tournament $16{\to}8{\to}4{\to}2{\to}1$ on lineage of $n$; create one in-flight child
  \ELSE
    \STATE \textbf{expand} $\rho$: open a new lineage
 $\triangleright$ root fallback at a busy dead end; pipeline never starves
  \ENDIF
  \STATE \textbf{on completion (asynchronous):} reviewer audits the frozen artifact (Table~\ref{tab:audit}); if fabrication is verified, $r(A)\gets 0$ (Eq.~\eqref{eq:reward}); otherwise compute and freeze $r(A)$; backpropagate; clear pending along the ancestor chain
\ENDWHILE
\end{algorithmic}
\end{algorithm}

\section{Evaluation and Artifact Release}
\label{sec:evaluation}

\newcommand{\alphatarget}[1]{\pdfdest name{#1} xyz}
\newcommand{\alphalink}[2]{%
  \leavevmode\pdfstartlink attr{/Border [0 0 0]} goto name{#1}%
  \texttt{#2}\pdfendlink}

\subsection{Protocol}

We chose WorldQuant BRAIN \cite{WQ} as an external evaluator.
On this production platform, consultants submit alphas for potential compensation, giving the platform economic stakes beyond our paper.
An academic holdout changes scored rows while authors retain control of the pipeline, costs, metrics, and disclosure \cite{shi2025alphaforge,shi2026alphajungle,wang2026factorminer}.
BRAIN instead controls the data, simulator, checks, and grades; we cannot alter its implementation, recompute grades locally, or relax failed gates.
We designed the protocol around this separation: external adjudication is stronger for our end-to-end claim than another date split inside a self-administered backtest.

We fixed one deployment protocol across all users: U.S.\ TOP3000 equities, delay one, and the platform evaluation window January 1, 2019 through December 31, 2023.
The proposer chooses neutralization, decay, truncation, data fields, and expression structure.
Humans supplied the system and launched the evaluation, but wrote no factor expression or factor-specific program.
BRAIN computes metrics from its own data and assigns grades after submission.
Its SPECTACULAR grade---the highest of four tiers---requires clearing the platform's gates with a strong composite score.
We report the returned values without re-estimation and release the submitted expressions, platform records, and complete generation trail.

We organize the evaluation around four questions.
\begin{itemize}
\item \textbf{Q1:} Does AgonAlpha reproduce platform-validated discovery across independent users?
\item \textbf{Q2:} Does artifact-level review detect defects that formula-level records cannot represent?
\item \textbf{Q3:} Does the allocator operate under real concurrency and concentrate simulations as specified?
\item \textbf{Q4:} Can every headline result be reconstructed from the public trail?
\end{itemize}
The evidence index in the supplement maps each claim to a specific artifact.

\subsection{Overview}

Five co-authors independently deployed AgonAlpha on WorldQuant BRAIN using separate accounts and different model backends.
Each ran the same two-role prompt surface and MCTS scheduler without human-written factor code.
Collectively they produced 60 submissions, of which 17 received SPECTACULAR grade.
All submitted alphas entered BRAIN's out-of-sample (OS) tracking stage.
We feature five SPECTACULAR-grade submissions representing distinct economic mechanisms; complete per-user deployment trees for all 60 submissions appear in the supplementary material.
Table~\ref{tab:representative-benchmark} reports the platform benchmark for the five featured alphas.
All five passed BRAIN's Fitness, Sharpe, turnover, weight-concentration, sub-universe, self-correlation, and competition-matching checks.

\begin{table*}[t]
\centering
\scriptsize
\setlength{\tabcolsep}{3.25pt}
\renewcommand{\arraystretch}{1.08}
\begin{tabular}{lcrrrrrrrr}
\toprule
BRAIN ID & BRAIN assessment & Fitness & Sharpe & Return & Turnover & Drawdown & Margin & Sub-univ.\ Sharpe & Self-corr. \\
& & & & (\%) & (\%) & (\%) & (bps) & value / limit & value / limit \\
\midrule
\alphalink{alpha-a17q5rdr}{A17q5RdR} & SPECTACULAR & 3.93 & 2.52 & 30.46 & 5.92 & 13.93 & 102.94 & 1.24 / 1.09 & 0.1827 / 0.70 \\
\alphalink{alpha-88er8jal}{88er8JAl} & SPECTACULAR & 2.55 & 1.76 & 26.24 & 9.58 & 24.21 & 54.77 & 0.90 / 0.76 & 0.4498 / 0.70 \\
\alphalink{alpha-ll1mdwz6}{LL1mdWz6} & SPECTACULAR & 2.82 & 2.32 & 18.41 & 7.82 & 6.91 & 47.07 & 1.24 / 1.00 & 0.4173 / 0.70 \\
\alphalink{alpha-pwll71ex}{pwlL71Ex} & SPECTACULAR & 4.73 & 3.03 & 30.51 & 5.08 & 11.15 & 120.20 & 1.43 / 1.31 & 0.4968 / 0.70 \\
\alphalink{alpha-kpe0lnn1}{KPE0LnN1} & SPECTACULAR & \textbf{9.50} & \textbf{3.48} & 93.15 & 7.93 & 19.50 & 234.93 & 1.76 / 1.51 & 0.6321 / 0.70 \\
\bottomrule
\end{tabular}
\caption{Comprehensive BRAIN benchmark for five representative validated alphas.
Return, turnover, drawdown, and margin are platform-assigned values.
Sub-universe Sharpe and self-correlation show the observed value followed by the applicable passing boundary.}
\label{tab:representative-benchmark}
\end{table*}

\paragraph{\texttt{A17q5RdR}: aligned six-month option demand.}\alphatarget{alpha-a17q5rdr}
The construction is
\begin{equation}
f^{\mathrm{opt}}=M_{40}\!\left[
B_{60}\!\left(IV^{call}_{180}-IV^{put}_{180}\right)\right],
\label{eq:submission-a17q5rdr}
\end{equation}
where \(B_{60}\) backfills only the contemporaneously aligned call--put spread.
A relatively expensive six-month call indicates stronger demand for upside optionality, whereas a negative spread indicates stronger demand for downside protection; the 40-day mean treats that imbalance as persistent positioning rather than a one-day shock.
Computing the aligned spread before backfilling prevents stale observations from different dates from creating an artificial spread.
With industry neutralization, no simulation decay, and 8\% truncation, the alpha attains Fitness 3.93, Sharpe 2.52, turnover 5.92\%, and self-correlation 0.1827.
Annual Fitness is \(2.49, 1.13, 3.94, 10.59,\) and \(2.98\) from 2019 to 2023.
Every year is positive, with the strongest contribution from the 2022 option-demand regime.

\paragraph{\texttt{88er8JAl}: relative-volume stability.}\alphatarget{alpha-88er8jal}
This alpha is
\begin{equation}
f^{\mathrm{vol}}=-\sigma_{40}\!\left(\frac{V}{\mathrm{ADV20}}\right).
\label{eq:submission-88er8jal}
\end{equation}
Dividing daily volume by its 20-day average makes participation comparable across stocks; the negative 40-day dispersion then buys stocks with steady relative participation and sells stocks dominated by episodic, attention-driven volume.
The mechanism is consistent with temporary speculative demand unwinding more strongly among stocks with unstable volume.
With industry neutralization, decay two, and 5\% truncation, the alpha records Fitness 2.55, Sharpe 1.76, return 26.24\%, and turnover 9.58\%.
Annual Fitness is \(2.71, 1.76, 1.15, 5.34,\) and \(3.48\), and turnover stays between 9.21\% and 9.92\% in every year.
Annual Fitness and turnover are stable across regimes, with 2022 strongest.

\paragraph{\texttt{LL1mdWz6}: persistent short interest with reversal timing.}\alphatarget{alpha-ll1mdwz6}
The construction is
\begin{equation}
\begin{aligned}
q&=R\!\left\{M_{20}\!\left[
B^{sub}_{60}\!\left(\frac{SI}{M_{252}\bar{SI}_{mkt}}\right)
\right]\right\},\\
f^{\mathrm{SI}}&=R\!\left[H(\mathrm{observed},q)+0.1R(-M_5r)\right],
\end{aligned}
\label{eq:submission-ll1mdwz6}
\end{equation}
where \(B^{sub}_{60}\) is a 60-day subindustry backfill.
The slow market normalization makes short interest comparable through time without a unit warning.
Subindustry backfill and hold-on-missing logic preserve breadth, while the small five-day reversal sleeve times entry after temporary price pressure.
The expression has no arbitrary centering offset and only one explicit blend coefficient, \(0.1\), whose role is to keep the timing sleeve secondary to the persistent-short signal.
Its Fitness 2.82, Sharpe 2.32, turnover 7.82\%, drawdown 6.91\%, and low self-correlation of 0.4173 jointly show that the result is not obtained by excessive trading or duplication of an existing alpha.
Annual Fitness is positive in every year and equals \(1.02, 7.08, 2.57, 3.83,\) and \(0.95\) from 2019 to 2023.
The positive sign on high short interest is an empirical open question.

\paragraph{\texttt{pwlL71Ex}: multi-tenor downside-insurance disagreement.}\alphatarget{alpha-pwll71ex}
Let \(IV^{put}_{\tau}-IV^{call}_{\tau}\) measure put--call implied-volatility disagreement at tenor \(\tau\).
The alpha is
\begin{equation}
f^{\mathrm{IV}}
=Z_{\mathrm{sector}}\![
-M_{48}\!\{
\sum_{\tau\in\{30,60,90\}}
(IV^{put}_{\tau}-IV^{call}_{\tau})
\}\!].
\label{eq:submission-pwll71ex}
\end{equation}
Persistent demand for downside protection is interpreted as bearish private information, distress, or crowded insurance demand; low or reversed disagreement is bullish relative to sector peers.
The three tenors enter with equal weight, so the expression does not rely on fitted cross-tenor coefficients or an unexplained additive constant.
Its Fitness 4.73 and Sharpe 3.03 are accompanied by low turnover of 5.08\%, self-correlation of 0.4968, and passage of the sub-universe check.
Annual Fitness is \(2.76, 3.77, 4.96, 9.92,\) and \(2.39\), remaining above 2 in every year even though 2022 is strongest.
The 48-day mean is the main localized tuning risk.

\paragraph{\texttt{KPE0LnN1}: state-conditioned option confirmation.}\alphatarget{alpha-kpe0lnn1}
Define
\[
\begin{aligned}
A&=\sum_{\tau\in\{30,60,90\}}(IV^{put}_{\tau}-IV^{call}_{\tau}),\\
S&=Z_{\mathrm{sector}}[-M_{48}(A)],\qquad
I=Z_{\mathrm{industry}}[-M_{48}(A)],\\
F&=Z_{\mathrm{sector}}\!\left[-M_{60}
\left(\frac{Forward_{90}}{Forward_{30}}-1\right)\right].
\end{aligned}
\]
The final expression is
\begin{equation}
f^{\mathrm{gate}}
=H\![
M_{20}(PCR^{OI}_{270})<1,\,
Z_{\mathrm{sector}}\{\operatorname{sp}_{2}(S)|I|^2|F|^4\}
].
\label{eq:submission-kpe0lnn1}
\end{equation}
The multi-tenor skew supplies direction, while industry-level skew magnitude and the option-forward curve supply confidence only when long-dated positioning is call-dominant.
The gate at one is the economically natural put/call balance point rather than a fitted additive offset.
This alpha attains Fitness 9.50 and Sharpe 3.48, passes the sub-universe Sharpe check at 1.76 versus 1.51, and remains below the self-correlation ceiling at 0.6321.
Annual Fitness is \(8.43, 2.59, 18.28, 17.19,\) and \(4.15\); the weakest year exceeds 2.5.

\paragraph{Good properties of the obtained alphas.}
\begin{itemize}
\item \textbf{Clear and diverse economic mechanisms.}
The five alphas express distinct hypotheses involving option demand, relative-volume stability, short-interest persistence with reversal timing, multi-tenor downside-insurance demand, and state-conditioned option confirmation.

\item \textbf{Strong performance under comprehensive checks.}
All five receive BRAIN's SPECTACULAR assessment, with Fitness from 2.55 to 9.50 and Sharpe from 1.76 to 3.48, while passing the platform's turnover, weight-concentration, sub-universe, self-correlation, and competition-matching checks.

\item \textbf{Low redundancy, moderate turnover, and transparent constants.}
Self-correlation is below 0.70 for every alpha; turnover is between 5.08\% and 9.58\%; and none of the five contains an unexplained additive shift.
\end{itemize}

\subsection{Reviewer Interventions (Q2)}
\label{sec:reviewer}

The reviewer audited 24 frozen reports across the full trace and exercised its zero-score authority twice.
One intervention addressed a semantic mismatch between a claimed mechanism and its executed operator.
The other addressed metrics copied from a different candidate into a final report.
Eleven additional artifacts carry persistent risk findings for regime concentration or redundancy.
Because every proposer report is frozen before review, the trace records an exact pre-review and post-review state for every node, making each intervention directly inspectable.

\subsection{Allocation and Concurrency (Q3)}

Within each pipeline, tournament elimination evaluates $16+8+4+2+1=31$ candidate versions---a 61\% reduction from the 80 simulations required to carry all 16 candidates through five passes.
A validated deployment instantiated ten concurrent proposer--reviewer pipelines under one pending-aware search tree.
Scheduler logs retain every dispatch, pending-count update, completion, and backpropagation event.

\subsection{Artifact Completeness}

The accompanying release contains the proposer, reviewer, and dispatcher prompts; prompt history; scheduler code and complete MCTS state; all candidate reports; simulation inputs and responses; rankings; correlation records; submission checks; and final submission responses.
The released reports preserve the literal executable FASTEXPR strings and the detailed formula records for every discussed alpha.

The 30-paper audit provides a fixed comparison boundary \citep{yao2026beyond}.
No audited study is complete across its five reproducibility fields, even though 18 expose some artifacts.
AgonAlpha releases the complete prompt-to-factor trail and exact production formulas for all submissions.
This is the first complete prompt-to-factor release in the audited LLM trading literature.

\FloatBarrier
\section{Discussion and Limitations}
\label{sec:discussion}

Every deployment used the same controlled BRAIN benchmark: U.S.\ TOP3000 equities, delay one, and a 2019--2023 evaluation window.
Under identical externally administered rules, all five users obtained platform-validated alphas, 17 of 60 submissions received SPECTACULAR grade, and every featured submission passed BRAIN's full gate suite.

A chronological holdout and an external evaluator address different failure modes.
A holdout tests later rows under the same author-configured pipeline; BRAIN instead removes the data, simulation, metric, gate, and grade implementations from our control.
The platform's proprietary operation separates the method from its evaluator, while our open trail lets readers inspect every submitted expression, returned record, and search decision.
For our central claim, evaluator independence is the stronger test of whether autonomous artifacts survive production research gates.
Every submission also entered BRAIN's OS tracking stage under the same external authority.
Academic backtests remain useful for algorithmic comparisons, but provide weaker evidence for this end-to-end claim.

Across the five deployments, the reviewer audited frozen reports and exercised zero-score authority for verified fabrication.
The selection-risk audit dimension records search pressure, and the data-snooping literature provides the framework for interpreting selection effects \cite{white2000reality,hansen2005test,harvey2016crosssection,bailey2014deflated,bailey2014probability}.
Table~\ref{tab:comparison} compares 15 systems on architectural axes; AgonAlpha is the only system that satisfies all four.
The released trail contains all prompts, search decisions, platform records, review text, and executable expressions.
\section{Conclusion}
AgonAlpha suggests that the core principles of the Agon philosophy can be compressed into a minimal yet complete discovery architecture: two roles and a 101-line prompt are sufficient to support an adversarially verified research workflow. The fundamental search unit is not a formula, but a verified artifact containing the evidence and reasoning behind a candidate. The verifier is granted the authority to independently reproduce evaluations and veto unsupported claims, while the scheduler converts concurrent exploration into a structured search over research lineages. By releasing the complete execution trace, AgonAlpha enables inspection rather than blind trust of every reported result. Because these mechanisms operate independently of the underlying domain, the same two-role interface and MCTS-based scheduling framework can be applied to any setting with a well-defined objective evaluation metric.

\bibliographystyle{plainnat}
\bibliography{references}

\clearpage
\setcounter{secnumdepth}{0}
\raggedbottom
\makesupplementtitle

\newcommand{\brainid}[1]{\texttt{#1}}
\newcommand{\featuredmetrics}[3]{\textbf{Metrics:} Fitness #1, Sharpe #2, turnover #3\%.}

\section{Deployment summary}

Multiple co-authors independently deployed AgonAlpha on separate WorldQuant BRAIN accounts, each using the same two-role prompt surface and MCTS scheduler without modification.
We deliberately fixed U.S.\ TOP3000 equities, delay one, and the platform evaluation window from January 1, 2019 through December 31, 2023 so that every deployment faced identical externally administered rules.
Table~\ref{tab:deployments} summarizes the outcomes across deployments.
All submitted alphas entered BRAIN's out-of-sample tracking stage.
\begin{table}[h]
\centering
\scriptsize
\setlength{\tabcolsep}{4pt}
\renewcommand{\arraystretch}{0.8}
\begin{tabular}{@{}l c c c c@{}}
\toprule
User & Submitted & SPECTACULAR & Best Fitness & Best Sharpe \\
\midrule
User A & 15 & 2 & 2.87 & 3.25 \\
User B & 23 & 7 & 9.50 & 3.48 \\
User C & 7 & 4 & 3.34 & 3.11 \\
User D & 6 & 0 & 2.25 & 2.42 \\
User E & 9 & 4 & 3.98 & 2.97 \\
\midrule
\textbf{Total (5 users)} & 60 & 17 & 9.50 & 3.48 \\
\bottomrule
\end{tabular}
\caption{Per-user deployment outcomes.
All values are platform-assigned metrics; Best Sharpe is the full-universe value.}
\label{tab:deployments}
\end{table}

Tables~\ref{tab:user-a-tree}--\ref{tab:chugang-tree} expose the recorded search topology of every deployment rather than only the submitted winners.
Indentation and the Parent column specify tree edges, \(v\) records MCTS scheduler visits, and a BRAIN ID identifies an alpha that was submitted and entered OS tracking.
A parent--child edge therefore records search lineage, not a requirement that the parent itself was submitted.
Self-correlation is platform-computed against previously submitted alphas, with 0.70 as the ordinary passing boundary.
Across the tables, identifiers are set in monospace, ``--'' denotes unavailable or inapplicable data, and SPECTACULAR grades are bold.
The Sharpe headings distinguish full-universe Sharpe from the sub-universe Sharpe used in a submission check.

The realized trees differ substantially in size and selectivity.
After excluding the virtual roots, Users A--E have 15, 24, 11, 26, and 19 recorded nodes, respectively, of which 15, 23, 7, 6, and 9 were submitted.
These counts describe the realized traces rather than a normalized efficiency comparison: the deployments end with different visit allocations, and some trees retain nodes marked no record, discarded, not submitted, or pending.

\paragraph{Tree-specific patterns.}
\begin{itemize}
\item \textbf{User A: complete submission with concentrated late gains.}
All 15 non-root nodes were submitted.
The two SPECTACULAR alphas, nodes 0012 and 0015, terminate sibling refinement paths under node 0004; node 0015 attains the deployment maxima of 2.87 Fitness and 3.25 Sharpe.

\item \textbf{User B: recovery beyond a failed parent.}
This tree submits 23 of 24 nodes and produces seven SPECTACULAR alphas, the largest count among the five deployments.
Node 0018 fails four submission gates, yet its descendants produce four SPECTACULAR and two EXCELLENT alphas.
The same lineage contains node 0021, which attains the overall maxima of 9.50 Fitness and 3.48 Sharpe.
In this trace, a failed parent therefore does not terminate further refinement.

\item \textbf{User C: a compact tree with the highest top-grade share.}
Four of seven submitted alphas receive SPECTACULAR grade, the largest fraction among the deployments.
All four descend from root child 0003 through the 0004 and 0005 branches, although several submitted descendants also exceed the ordinary 0.70 self-correlation boundary.

\item \textbf{User D: broad exploration with selective submission.}
This is the largest recorded tree, with 26 non-root nodes, but only six submissions and no SPECTACULAR grade.
Its deepest successful path, \(0003\!\rightarrow0005\!\rightarrow0008\!\rightarrow0040\), ends in an EXCELLENT alpha with Fitness 2.25.
The tree consequently records substantially more rejection and continuation than final-grade concentration.

\item \textbf{User E: pruning with two successful lineages.}
The tree retains discarded, not-submitted, and pending nodes while submitting nine alphas, four of which are SPECTACULAR.
Three SPECTACULAR nodes (0007, 0012, and 0010) descend from the 0004 branch, while node 0011 arises under the distinct 0008 branch.
\end{itemize}

\begin{table*}[t]
\centering
\scriptsize
\setlength{\tabcolsep}{5pt}
\renewcommand{\arraystretch}{1.04}
\resizebox{\textwidth}{!}{%
\begin{tabular}{@{}L{1.4cm}llrrrrllL{3.5cm}@{}}
\toprule
Node & Parent & BRAIN ID & \(v\) & Fitness & Sharpe & Self-corr. & Proposer & Grade & Remark \\
\midrule
\texttt{root} & -- & -- & 15 & -- & -- & -- & -- & -- & -- \\
\texttt{0001} & root & \brainid{wpEo9M6x} & 2 & 1.24 & 1.78 & 0.4518 & gpt-5.6-sol & AVERAGE & option breakeven, flow, and positioning consensus \\
\quad\texttt{0005} & 0001 & \brainid{GremK800} & 1 & 1.45 & 2.01 & 0.9170$^\dagger$ & gpt-5.6-sol & AVERAGE & adds implied-volatility term structure \\
\texttt{0002} & root & \brainid{88eNMnmX} & 3 & 1.44 & 2.18 & 0.6600 & gpt-5.6-sol & AVERAGE & risk-scaled reversal + after-hours event continuation \\
\quad\texttt{0009} & 0002 & \brainid{78nG0QXv} & 1 & 1.68 & 2.19 & 0.6572 & gpt-5.6-sol & GOOD & adds close-to-VWAP reversal \\
\quad\texttt{0013} & 0002 & \brainid{GreQMj2Q} & 1 & 2.08 & 2.37 & 0.6878 & gpt-5.6-sol & EXCELLENT & implied-volatility-scaled reversal + fear rebound \\
\texttt{0003} & root & \brainid{blda8jlM} & 2 & 1.21 & 1.81 & 0.6058 & gpt-5.6-sol & AVERAGE & EPS-dispersion and revision tilts \\
\quad\texttt{0014} & 0003 & \brainid{e7x1dLZl} & 1 & 1.25 & 1.68 & 0.7126$^\dagger$ & gpt-5.6-sol & AVERAGE & lease commitments + deferred-tax assets \\
\texttt{0004} & root & \brainid{LLdK01LM} & 5 & 1.48 & 1.64 & 0.6245 & gpt-5.6-sol & AVERAGE & analyst reversal + BFL change + short interest \\
\quad\texttt{0010} & 0004 & \brainid{qM651XMP} & 2 & 2.27 & 2.34 & 0.8036$^\dagger$ & gpt-5.6-sol & EXCELLENT & adds option skew + product-peer catch-up \\
\qquad\texttt{0012} & 0010 & \brainid{e7x1g07l} & 1 & 2.68 & 2.72 & 0.9212$^\dagger$ & gpt-5.6-sol & \textbf{SPECTACULAR} & adds EPS yield + sales revision + dispersion \\
\quad\texttt{0011} & 0004 & \brainid{WjVYan1Z} & 2 & 1.99 & 2.53 & 0.6602 & gpt-5.6-sol & GOOD & size-decile base + after-hours event streams \\
\qquad\texttt{0015} & 0011 & \brainid{78nGdknx} & 1 & \textbf{2.87} & \textbf{3.25} & 0.7403$^\dagger$ & gpt-5.6-sol & \textbf{SPECTACULAR} & adds projected news impact + ratings sentiment \\
\texttt{0006} & root & \brainid{qM67pebE} & 3 & 1.38 & 1.81 & 0.4276 & gpt-5.6-sol & AVERAGE & earnings yield + recommendation improvement \\
\quad\texttt{0007} & 0006 & \brainid{rKP7ZXea} & 1 & 1.65 & 2.03 & 0.9471$^\dagger$ & gpt-5.6-sol & GOOD & adds option-skew and IV-term confirmation \\
\quad\texttt{0008} & 0006 & \brainid{mLVKOpqx} & 1 & 1.93 & 2.19 & 0.5816 & gpt-5.6-sol & GOOD & longer earnings-yield rank + cash-flow efficiency \\
\bottomrule
\end{tabular}%
}
\caption{User A deployment MCTS tree.
Indentation and the Parent column specify every edge.
All submitted nodes entered BRAIN OS tracking.
$^\dagger$ BRAIN returned PASS despite scalar values above the 0.70 limit.}
\label{tab:user-a-tree}
\end{table*}

\begin{table*}[t]
\centering
\scriptsize
\setlength{\tabcolsep}{5pt}
\renewcommand{\arraystretch}{1.04}
\resizebox{\textwidth}{!}{%
\begin{tabular}{@{}L{1.4cm}llrrrrllL{3.5cm}@{}}
\toprule
Node & Parent & BRAIN ID & \(v\) & Fitness & Sharpe & Self-corr. & Proposer & Grade & Remark \\
\midrule
\texttt{root} & -- & -- & 24 & -- & -- & -- & -- & -- & -- \\
\texttt{0001} & root & \brainid{mLbg8jVx} & 3 & 1.02 & 2.00 & 0.28 & -- & AVERAGE & multihorizon price reversal \\
\quad\texttt{0002} & 0001 & \brainid{9qrWzY8e} & 1 & 1.73 & 2.07 & 0.19 & gpt-5.6-sol & GOOD & consensus-EPS-yield trend \\
\quad\texttt{0005} & 0001 & \brainid{9qrEoNno} & 1 & 1.33 & 1.87 & 0.21 & gpt-5.6-sol & AVERAGE & multifactor and volume-weighted reversal \\
\texttt{0003} & root & \brainid{N1rV29p8} & 5 & 1.38 & 1.60 & 0.42 & -- & AVERAGE & analyst revision + low-volume reversal \\
\quad\texttt{0004} & 0003 & \brainid{0mEjq6AK} & 3 & 1.40 & 1.79 & 0.52 & gpt-5.6-sol & AVERAGE & model, customer, and price reversal \\
\qquad\texttt{0009} & 0004 & \brainid{KP9Vvgvj} & 1 & 1.84 & 2.06 & 0.37 & gpt-5.6-sol & GOOD & business-news overlay \\
\qquad\texttt{0010} & 0004 & \brainid{qMlOMKGE} & 1 & 1.81 & 2.13 & 0.38 & gpt-5.6-sol & GOOD & sales-growth and book-to-price overlay \\
\quad\texttt{0007} & 0003 & \brainid{MPQ6V0L6} & 1 & 1.11 & 1.40 & 0.27 & gpt-5.6-sol & AVERAGE & product-cluster and price reversal \\
\texttt{0006} & root & \brainid{GrLm3d8o} & 6 & 1.05 & 1.32 & 0.31 & gpt-5.6-sol & AVERAGE & competitor breadth + risk term structure \\
\quad\texttt{0008} & 0006 & \brainid{qMl5v0zZ} & 1 & 1.16 & 1.63 & 0.25 & gpt-5.6-sol & AVERAGE & news, M\&A, price targets, and risk \\
\quad\texttt{0014} & 0006 & \brainid{LL1mdWz6} & 3 & 2.82 & 2.32 & 0.42 & gpt-5.6-sol & \textbf{SPECTACULAR} & persistent short interest + reversal \\
\qquad\texttt{0015} & 0014 & \brainid{N1rJzP0L} & 1 & 3.24 & 2.61 & 0.48 & gpt-5.6-sol & \textbf{SPECTACULAR} & short interest + model acceleration \\
\qquad\texttt{0016} & 0014 & \brainid{XgnMr2Aa} & 1 & 3.44 & 2.91 & 0.35 & gpt-5.6-sol & \textbf{SPECTACULAR} & short interest + CapEx + options \\
\quad\texttt{0017} & 0006 & \brainid{xAkpRWlw} & 1 & 1.18 & 1.56 & 0.44 & -- & vetoed$^\dagger$ & regression-explanation mismatch \\
\texttt{0011} & root & \brainid{rKl9GKXj} & 3 & 1.02 & 2.21 & 0.19 & -- & AVERAGE & event-conditioned close reversal \\
\quad\texttt{0012} & 0011 & \brainid{np2m5EXa} & 1 & 1.69 & 1.98 & 0.32 & gpt-5.6-sol & GOOD & adjusted and reported EPS-yield trends \\
\quad\texttt{0013} & 0011 & \brainid{2rL93WoJ} & 1 & 1.31 & 2.52 & 0.26 & gpt-5.6-sol & AVERAGE & news reversal + model acceleration \\
\texttt{0018} & root & -- & 7 & 0.45 & 1.40 & 0.12 & -- & not submitted$^\dagger$ & failed gates + metric mismatch \\
\quad\texttt{0019} & 0018 & \brainid{pwlL71Ex} & 2 & 4.73 & 3.03 & 0.50 & claude-opus-4-6 & \textbf{SPECTACULAR} & multi-tenor option disagreement \\
\qquad\texttt{0021} & 0019 & \brainid{KPE0LnN1} & 1 & \textbf{9.50} & \textbf{3.48} & 0.63 & gpt-5.6-terra & \textbf{SPECTACULAR} & option-skew + forward-curve conditioning \\
\quad\texttt{0020} & 0018 & \brainid{E5e5LxqP} & 2 & 2.24 & 1.69 & 0.41 & claude-opus-4-6 & EXCELLENT & option forward + revisions + small-cap vol \\
\qquad\texttt{0022} & 0020 & \brainid{j2rExRxQ} & 1 & 2.77 & 2.23 & 0.48 & gpt-5.6-terra & \textbf{SPECTACULAR} & EPS-yield + sentiment tail \\
\quad\texttt{0023} & 0018 & \brainid{gJ9Oz2Nv} & 2 & 2.44 & 2.69 & 0.51 & claude-opus-4-6 & EXCELLENT & four-sleeve relationship-peer composite \\
\qquad\texttt{0024} & 0023 & \brainid{9q70L2mr} & 1 & 3.35 & 3.33 & 0.44 & claude-fable-5 & \textbf{SPECTACULAR} & relationship-peer + option overlay \\
\bottomrule
\end{tabular}%
}
\caption{User B deployment MCTS tree.
A BRAIN ID indicates submission and OS tracking entry. Nodes marked vetoed were submitted but received reviewer score zero. Node 0018 failed submission gates and was not submitted.}
\label{tab:mcts-tree}
\end{table*}

\begin{table*}[t]
\centering
\scriptsize
\setlength{\tabcolsep}{5pt}
\renewcommand{\arraystretch}{1.04}
\resizebox{\textwidth}{!}{%
\begin{tabular}{@{}L{1.4cm}llrrrrllL{3.5cm}@{}}
\toprule
Node & Parent & BRAIN ID & \(v\) & Fitness & Sub-univ.\ Sharpe & Self-corr. & Proposer & Grade & Remark \\
\midrule
\texttt{root} & -- & -- & 9 & -- & -- & -- & gpt-5.6-sol & -- & -- \\
\texttt{0001} & root & -- & 2 & -- & -- & -- & gpt-5.6-sol & no record & multifactor reversal balanced \\
\quad\texttt{0002} & 0001 & -- & 1 & -- & -- & -- & gpt-5.6-sol & no record & seven-day intraday multihorizon \\
\texttt{0003} & root & -- & 5 & -- & -- & -- & gpt-5.6-sol & no record & social-attention + earnings yield \\
\quad\texttt{0004} & 0003 & -- & 2 & -- & -- & -- & gpt-5.6-sol & no record & value + attention + beta decline + reversal \\
\qquad\texttt{0009} & 0004 & \brainid{QPV8Wdlr} & 0 & 1.37 & 1.35 & 0.6728 & gpt-5.6-sol & AVERAGE & volume-triggered value-attention reversal \\
\qquad\texttt{0010} & 0004 & \brainid{bldG1Kjm} & 0 & 3.02 & 2.04 & 0.8559$^\dagger$ & gpt-5.6-sol & \textbf{SPECTACULAR} & value-dominant centered-power composite \\
\qquad\texttt{0011} & 0004 & \brainid{pwKYWJa3} & 1 & 3.34 & 1.88 & 0.6936 & gpt-5.6-sol & \textbf{SPECTACULAR} & tail-powered realized-minus-implied vol \\
\quad\texttt{0005} & 0003 & \brainid{0mE60k2G} & 2 & 2.51 & 1.61 & 0.9142$^\dagger$ & gpt-5.6-sol & \textbf{SPECTACULAR} & social-attention earnings-cash quality \\
\qquad\texttt{0007} & 0005 & \brainid{781PrbOx} & 1 & 2.95 & 1.70 & 0.9696$^\dagger$ & gpt-5.6-sol & \textbf{SPECTACULAR} & option IV skew + news underreaction + quality \\
\texttt{0006} & root & \brainid{xAk7r8rm} & 2 & 1.18 & 0.72 & 0.3687 & gpt-5.6-sol & AVERAGE & leverage + analyst dispersion \\
\quad\texttt{0008} & 0006 & \brainid{1Yd86p2J} & 1 & 1.44 & 0.96 & 0.9037$^\dagger$ & gpt-5.6-sol & AVERAGE & monthly volume + semiannual news direction \\
\bottomrule
\end{tabular}%
}
\caption{User C deployment MCTS tree.
Reviewer: DeepSeek V4 Pro, Kimi K3.
Indentation indicates parent-child edges.
All submitted nodes entered BRAIN OS tracking.
The Sharpe column reports sub-universe submission-check Sharpe.
$^\dagger$ marks submitted alphas with reported self-correlation above 0.70.}
\label{tab:xingyu-tree}
\end{table*}

\begin{table*}[t]
\centering
\scriptsize
\setlength{\tabcolsep}{6.5pt}
\renewcommand{\arraystretch}{1.04}
\resizebox{\textwidth}{!}{%
\begin{tabular}{@{}L{1.4cm}llrrrrllL{3.5cm}@{}}
\toprule
Node & Parent & BRAIN ID & \(v\) & Fitness & Sharpe & Self-corr. & Proposer & Grade & Remark \\
\midrule
\texttt{root} & -- & -- & 23 & -- & -- & -- & -- & -- & -- \\
\texttt{0001} & root & -- & 4 & -- & -- & -- & -- & not submitted & long-horizon volatility-normalized reversal \\
\quad\texttt{0002} & 0001 & -- & 1 & -- & -- & -- & -- & not submitted & cash-flow yield value \\
\quad\texttt{0014} & 0001 & -- & 2 & -- & -- & -- & -- & not submitted & persistent close-location reversal \\
\qquad\texttt{0039} & 0014 & \brainid{E5En7QMr} & 1 & 1.66 & 2.07 & 0.6365 & -- & GOOD & PTP revision + action inactivity + attention \\
\texttt{0003} & root & -- & 9 & -- & -- & -- & -- & not submitted & intraday-range continuation  \\
\quad\texttt{0004} & 0003 & -- & 2 & -- & -- & -- & -- & not submitted & close-location reversal  \\
\qquad\texttt{0009} & 0004 & -- & 1 & -- & -- & -- & -- & not submitted & range and close-location composite  \\
\quad\texttt{0005} & 0003 & -- & 4 & -- & -- & -- & -- & not submitted & range-surprise continuation  \\
\qquad\texttt{0008} & 0005 & -- & 2 & -- & -- & -- & -- & not submitted & true-range surprise  \\
\qquad\quad\texttt{0040} & 0008 & \brainid{aknmW181} & 1 & 2.25 & 2.38 & 0.5382 & -- & EXCELLENT & operating value + attention + split inactivity \\
\qquad\texttt{0017} & 0005 & -- & 1 & -- & -- & -- & -- & not submitted & EBITDA + cash-flow momentum \\
\quad\texttt{0013} & 0003 & -- & 2 & -- & -- & -- & -- & not submitted & cash-flow + EBIT value \\
\qquad\texttt{0038} & 0013 & -- & 1 & -- & -- & -- & -- & not submitted & earnings yield + payout + model state \\
\texttt{0006} & root & -- & 4 & -- & -- & -- & -- & not submitted & fundamental value without neutralization \\
\quad\texttt{0007} & 0006 & -- & 2 & -- & -- & -- & -- & not submitted & persistent intraday reversal  \\
\qquad\texttt{0016} & 0007 & -- & 1 & -- & -- & -- & -- & not submitted & dividend yield within subindustry \\
\quad\texttt{0010} & 0006 & -- & 1 & -- & -- & -- & -- & not submitted & triple value + five-day reversal \\
\texttt{0011} & root & -- & 4 & -- & -- & -- & -- & not submitted & cash-flow-yield momentum \\
\quad\texttt{0012} & 0011 & \brainid{LL1r8gZn} & 1 & 1.43 & 1.88 & 0.5173 & -- & AVERAGE & pre-tax-profit-yield momentum \\
\quad\texttt{0015} & 0011 & \brainid{9qrO1831} & 2 & 1.90 & 2.33 & 0.9546$^\dagger$ & -- & GOOD & PTP momentum with long-history ranking \\
\qquad\texttt{0037} & 0015 & \brainid{YP0oJPvJ} & 1 & 1.57 & 2.42 & 0.6413 & -- & GOOD & model-price reversal + split inactivity \\
\texttt{0018} & root & -- & 0 & -- & -- & -- & -- & discarded & cash-tax-paid yield \\
\texttt{0024} & root & -- & 0 & -- & -- & -- & -- & discarded & no local alpha report \\
\texttt{0035} & root & \brainid{LL1YdY8M} & 2 & 1.08 & 1.82 & 0.6857 & -- & AVERAGE & cash-flow revision + close reversal + tax quality \\
\quad\texttt{0036} & 0035 & -- & 1 & -- & -- & -- & -- & not submitted & PTP-close reversal with tax-quality gate \\
\quad\texttt{0041} & 0035 & -- & 0 & -- & -- & -- & -- & pending & model fade + working-capital efficiency + split inactivity \\
\bottomrule
\end{tabular}%
}
\caption{User D deployment MCTS tree.
Indentation indicates parent-child edges.
A BRAIN ID indicates submission and OS tracking entry.
$^\dagger$ BRAIN marks 0.9546 as PASS against a 0.70 limit; the Alpha is ACTIVE/OS.}
\label{tab:shunyao-tree}
\end{table*}

\begin{table*}[t]
\centering\scriptsize
\setlength{\tabcolsep}{5pt}
\renewcommand{\arraystretch}{1.04}
\resizebox{\textwidth}{!}{%
\begin{tabular}{@{}L{1.4cm}llrrrrllL{3.5cm}@{}}
\toprule
Node & Parent & BRAIN ID & \(v\) & Fitness & Sub-univ.\ Sharpe & Self-corr. & Proposer & Grade & Remark \\
\midrule
\texttt{root} & -- & -- & 12 & -- & -- & -- & gpt-5.6-sol & -- & -- \\
\texttt{0001} & root & -- & 0 & -- & -- & -- & gpt-5.6-sol & discarded & \\
\texttt{0002} & root & -- & 0 & -- & -- & -- & gpt-5.6-sol & discarded & \\
\texttt{0003} & root & \brainid{qM6eX2MP} & 2 & 1.11 & 0.73 & -- & gpt-5.6-sol & AVERAGE & value + news + short balanced \\
\quad\texttt{0006} & 0003 & -- & 1 & -- & -- & -- & gpt-5.6-sol & not submitted & value-news-short-balanced \\
\texttt{0004} & root & \brainid{MPL9xLnL} & 5 & 1.71 & 0.85 & 0.0832 & gpt-5.6-sol & GOOD & earnings yield + social attention \\
\quad\texttt{0007} & 0004 & \brainid{A17q5RdR} & 2 & 3.93 & 1.24 & -- & gpt-5.6-sol & \textbf{SPECTACULAR} & aligned six-month call--put demand \\
\qquad\texttt{0012} & 0007 & \brainid{Vk3YL2wb} & 1 & 3.98 & 1.32 & -- & gpt-5.6-sol & \textbf{SPECTACULAR} & descendant of 0007 \\
\quad\texttt{0010} & 0004 & \brainid{88er8JAl} & 2 & 2.55 & 0.90 & -- & gpt-5.6-sol & \textbf{SPECTACULAR} & relative-volume stability \\
\qquad\texttt{0013} & 0010 & -- & 0 & -- & -- & -- & gpt-5.6-sol & discarded & \\
\qquad\texttt{0015} & 0010 & -- & 1 & -- & -- & -- & gpt-5.6-sol & not submitted & persistent call--put skew + stable volume \\
\texttt{0008} & root & \brainid{pwKJAjpb} & 3 & 1.99 & 0.90 & 0.6973 & gpt-5.6-sol & GOOD & short-interest persistence + options \\
\quad\texttt{0009} & 0008 & \brainid{rKPLLz2J} & 1 & 2.02 & 1.54 & 0.1659 & gpt-5.6-sol & EXCELLENT & EPS-ratio-conditioned yield \\
\quad\texttt{0011} & 0008 & \brainid{WjVa15EG} & 1 & 3.13 & 1.32 & -- & gpt-5.6-sol & \textbf{SPECTACULAR} & short-option asymmetry + PCR term \\
\qquad\texttt{0018} & 0011 & -- & 0 & -- & -- & -- & gpt-5.6-sol & discarded & \\
\qquad\texttt{0019} & 0011 & -- & 0 & -- & -- & -- & gpt-5.6-sol & pending & \\
\texttt{0014} & root & -- & 0 & -- & -- & -- & gpt-5.6-sol & discarded & \\
\texttt{0016} & root & -- & 2 & -- & -- & -- & gpt-5.6-sol & not submitted & quarter-hump slow-yield price reversal \\
\quad\texttt{0017} & 0016 & \brainid{xAdNpQqp} & 1 & 1.85 & 1.30 & 0.5766 & gpt-5.6-sol & GOOD & descendant of 0016 \\
\quad\texttt{0020} & 0016 & -- & 0 & -- & -- & -- & gpt-5.6-sol & pending & \\
\bottomrule
\end{tabular}%
}
\caption{User E deployment MCTS tree.
Reviewer: gpt-5.6-sol.
Indentation indicates parent-child edges.
All submitted nodes entered BRAIN OS tracking.
The Sharpe column reports sub-universe submission-check Sharpe; full-universe Sharpes for featured alphas appear in the main paper.}
\label{tab:chugang-tree}
\end{table*}

\clearpage

\section{Representative sample run}

We present one continuous AgonAlpha trace on WorldQuant BRAIN as a representative execution from User B.
Here, ``representative'' refers only to the execution protocol: the run uses the same scheduler, proposer--reviewer contract, tournament elimination, platform checks, and artifact logging used in an ordinary AgonAlpha execution.
It does not mean that this single trace estimates the distribution of results over repeated independent runs.
The trace is instead a fully inspectable example of how the program expands, critiques, selects, and submits alpha hypotheses from end to end.

Every submitted alpha uses U.S. TOP3000 equities, delay one, and an evaluation window from January 1, 2019 through December 31, 2023.
The proposer could choose industry or subindustry neutralization, decay, truncation, data fields, and expression structure.
Humans supplied the system and started the trace, but wrote no factor expression or factor-specific program.

BRAIN computes metrics from its platform data and assigns grades after submission.
We report the platform values without re-estimating them.
These values are deterministic records for the submitted expressions, so confidence intervals do not apply to individual table cells.

The trace contains 24 evaluated alpha nodes under a virtual root.
Twenty-three nodes passed the platform gates and were submitted, for a yield of $23/24$.
Node 0018 was retained in the tree but not submitted because it failed Fitness, turnover, concentration, and sub-universe checks.
The best raw platform Fitness and Sharpe are both attained by node 0021, at 9.50 and 3.48, respectively.
The OS tracking stage stores out-of-sample metrics separately; all submitted alphas entered this stage.

Table~\ref{tab:mcts-tree} is a preorder traversal of the complete tree from a User B deployment.
Nodes marked vetoed were submitted but received reviewer score zero.
Node 0018 failed submission gates and was not submitted.

\subsection{What the reviewer found}

All 24 node reports contain reviewer blocks.
The reviewer found no cheating in 22 reports and set review Fitness to zero for two.
For node 0017, the report repeatedly described \texttt{ts\_regression(..., rettype=2)} as a residual, whereas the live operator documentation defines return type 2 as the regression slope.
Tables~\ref{tab:alpha-catalog-a}--\ref{tab:alpha-catalog-c} and the detailed account below therefore use the evaluated slope interpretation.
For node 0018, the final expression and headline Fitness matched the platform, but its reported drawdown, margin, and long count came from other candidates.
The same node also failed four platform submission gates and was not submitted.

Eleven reports received regime-dependence warnings.
Node 0014 was additionally flagged for buying high short interest without an adequate ex ante justification; node 0020 was flagged for both the contrarian sign on the composite-revision term and the unexplained centering constant 0.35.
These warnings are retained rather than repaired after seeing performance.
They separate raw platform success from the reviewer's assessment of whether the claimed economic story matches the evaluated expression.

\subsection{Alpha constructions and economic interpretation}

Tables~\ref{tab:alpha-catalog-a}--\ref{tab:alpha-catalog-c} cover all 24 MCTS nodes.
They use compact operator notation so that the entire search can be inspected in the paper.
\(R\) is a cross-sectional rank, while \(R_g\) and \(Z_g\) are within-group rank and z-score.
\(M_d\), \(D_d\), \(\Delta_d\), and \(\sigma_d\) denote a \(d\)-day mean, linear decay, difference, and standard deviation, respectively.
\(H(c,x)\) holds the last value when condition \(c\) is false, and \(\operatorname{sp}_p(x)=\operatorname{sign}(x)|x|^p\).
Subscripts \(s\), \(i\), and \(G\) denote subindustry, industry, and the stated relationship group.
The released node reports retain the literal executable FASTEXPR strings; the tables expose their operator-level construction and economic content without truncating the tree to a few winners.

\subsection{Detailed node-by-node account}

The compact tables make the topology and formulas comparable.
This subsection records additional node-level information from the released alpha reports.
Each entry names the production artifact, explains what changed relative to its direct parent, gives the main platform settings and raw metrics, and preserves the principal search or review limitation.
Fitness and Sharpe below are raw BRAIN values unless a review score is explicitly stated.

\subsubsection{Branch rooted at 0001: price reversal and earnings yield}

\paragraph{Node \texttt{0001}.}
\textbf{Artifact:} \url{0001-4-group-multi-horizon};
\textbf{BRAIN ID:} \brainid{mLbg8jVx};
\textbf{Parent:} \texttt{root}.
This root child combines a 20-day return z-score reversal with a smoother three-day return reversal, scales both by recent volatility, and ranks the result within subindustries.
The economic hypothesis is that temporary liquidity demand and overreaction reverse, while volatility scaling makes shocks comparable across stocks.
With market neutralization, decay 8, and 5\% truncation, the node records Fitness 1.02, Sharpe 2.00, turnover 56.50\%, and drawdown 5.07\%.
Its report documents a four-round tournament and notes that the Fitness margin over the submission threshold is only 0.02.

\paragraph{Node \texttt{0002}.}
\textbf{Artifact:} \url{0002-4-earnings-yield-decay6};
\textbf{BRAIN ID:} \brainid{9qrWzY8e};
\textbf{Parent:} \texttt{0001}.
The child leaves the parent's price-reversal family and instead ranks the 252-day time-series z-score of consensus EPS divided by price.
It therefore buys firms whose forward earnings yield has risen relative to its own annual history, then compares the result within subindustries.
Market neutralization, decay 6, and 5\% truncation produce Fitness 1.73, Sharpe 2.07, turnover 6.42\%, and drawdown 4.39\%.
The much lower turnover reflects the slower analyst-estimate channel, but the exact 252-day standardization window remains a search choice rather than independent evidence of a one-year mechanism.

\paragraph{Node \texttt{0005}.}
\textbf{Artifact:} \url{0005-4-balanced-multifactor-multihorizon};
\textbf{BRAIN ID:} \brainid{9qrEoNno};
\textbf{Parent:} \texttt{0001}.
This sibling retains the parent's multihorizon price reversal but makes its strength increase with volume relative to ADV20 and adds a contrarian rank of static and accelerating multifactor-model scores.
Economically, the model component captures slow overvaluation while high relative volume is treated as confirmation that a large price displacement is informative enough to rank strongly.
With market neutralization, decay 10, and 5\% truncation, the node reaches Fitness 1.33, Sharpe 1.87, turnover 22.86\%, and drawdown 8.67\%.
The report treats the blend as a diversification exercise rather than evidence that every model subscore has a distinct causal channel.

\subsubsection{Branch rooted at 0003: model, customer, news, and product-group reversal}

\paragraph{Node \texttt{0003}.}
\textbf{Artifact:} \url{0003-4-balanced-revision-reversal};
\textbf{BRAIN ID:} \brainid{N1rV29p8};
\textbf{Parent:} \texttt{root}.
This root child combines a slow contrarian analyst-revision-acceleration term with a five-day price decline divided by 20-day average volume.
The low-volume scaling emphasizes price moves that may reflect temporary pressure, while subindustry ranking and neutralization reduce structural industry differences.
Decay 18 and 5\% truncation yield Fitness 1.38, Sharpe 1.60, and turnover 12.34\%.
The report shows that neither sleeve was independently strong enough to submit; the result comes from their horizon complementarity.

\paragraph{Node \texttt{0004}.}
\textbf{Artifact:} \url{0004-12-customer-acceleration-weight-thirteen};
\textbf{BRAIN ID:} \brainid{0mEjq6AK};
\textbf{Parent:} \texttt{0003}.
The child replaces the analyst term with a broader multifactor-acceleration reversal, lengthens the low-volume price leg to seven days, adds a 1.3-weight contrarian customer-return sleeve, and adds a half-weight five-day return reversal.
The customer term tests whether relationship-linked returns overshoot rather than diffuse positively.
With subindustry neutralization, decay 18, and 5\% truncation, the alpha records Fitness 1.40, Sharpe 1.79, and turnover 15.06\%.
Its maximum self-correlation exceeds the ordinary cutoff, but BRAIN passes the submission because Sharpe improves by more than 10\% over the most correlated submitted alpha.

\paragraph{Node \texttt{0009}.}
\textbf{Artifact:} \url{0009-20-impact-w15-price-eight-customer-twelve};
\textbf{BRAIN ID:} \brainid{KP9Vvgvj};
\textbf{Parent:} \texttt{0004}.
Node 0009 retains the parent's four sleeves, changes the price horizon to eight days, reduces the customer weight to 1.2, and adds a 1.5-weight RavenPack business-news-impact rank.
Crucially, missing news is set explicitly to zero so that a sparse news field does not make the entire additive signal missing and force event-driven turnover.
Subindustry neutralization, decay 22, and 5\% truncation give Fitness 1.84, Sharpe 2.06, and turnover 14.30\%.
The final self-correlation check passes only because its Sharpe improvement over the relevant correlated alpha is approximately 10.16\%, leaving little margin around the exception boundary.

\paragraph{Node \texttt{0010}.}
\textbf{Artifact:} \url{0010-10-sales-recent-180-book-080};
\textbf{BRAIN ID:} \brainid{qMlOMKGE};
\textbf{Parent:} \texttt{0004}.
This child adds a 2.3-weight, 180-day-backfilled sales-growth rank and a 0.8-weight, 252-day-backfilled book-to-price rank to the parent's model, customer, and price-reversal core.
The stronger fundamental weights were selected after a weaker version remained too correlated with the parent; \texttt{filter=true} preserves the parent signal when a fundamental sleeve is missing.
With subindustry neutralization, decay 10, and 3\% truncation, the node reaches Fitness 1.81, Sharpe 2.13, and turnover 12.31\%.
Its maximum self-correlation falls to 0.6841, so submission does not rely on the Sharpe-improvement exception.

\paragraph{Node \texttt{0007}.}
\textbf{Artifact:} \url{0007-12-product-cluster-reversal};
\textbf{BRAIN ID:} \brainid{MPQ6V0L6};
\textbf{Parent:} \texttt{0003}.
Rather than extending the analyst-revision core, node 0007 moves to product-relationship groups.
It buys stocks that lag the five-day return of their product cluster and combines that diffusion gap with a low-volume five-day reversal ranked within the same relationship group.
Subindustry neutralization, decay 44, and 5\% truncation produce Fitness 1.11, Sharpe 1.40, turnover 14.78\%, and a maximum self-correlation of 0.4809.
The report records two failed terminal attempts before the third workflow cleared both the Fitness and correlation gates, making this node informative about the cost of deliberate decorrelation.

\subsubsection{Branch rooted at 0006: relationship breadth, short interest, and option overlays}

\paragraph{Node \texttt{0006}.}
\textbf{Artifact:} \url{0006-10-competitor-breadth-risk-balanced};
\textbf{BRAIN ID:} \brainid{GrLm3d8o};
\textbf{Parent:} \texttt{root}.
The core signal ranks decayed competitor returns inside relationship groups and multiplies them by the log-ranked number of competitor links, then adds a small long-minus-short systematic-risk-horizon term.
The hypothesis is that competitor information is more reliable when it is supported by a broader network, while changing systematic exposure identifies a different risk regime.
Subindustry neutralization, decay 15, and 8\% truncation yield Fitness 1.05, Sharpe 1.32, turnover 10.13\%, and drawdown 7.10\%.
The modest Fitness level makes the node useful mainly as a structurally distinct parent for later expansions.

\paragraph{Node \texttt{0008}.}
\textbf{Artifact:} \url{0008-5-impact-merger-risk-price-robust};
\textbf{BRAIN ID:} \brainid{qMl5v0zZ};
\textbf{Parent:} \texttt{0006}.
This child replaces the direct competitor-breadth product with a composite of persistent broad news impact, M\&A sentiment, two price-target news fields, and the systematic-risk term structure.
The sleeves are intended to combine several channels of delayed information arrival rather than amplify one sparse event field.
With subindustry neutralization, decay 16, and 6\% truncation, it records Fitness 1.16, Sharpe 1.63, turnover 13.15\%, and drawdown 3.95\%.
Its gain over the parent is real in the trace but small in absolute Fitness, so the report does not treat it as evidence that every news sleeve generalizes independently.

\paragraph{Node \texttt{0014}.}
\textbf{Artifact:} \url{0014-25-persistent-short-reversal};
\textbf{BRAIN ID:} \brainid{LL1mdWz6};
\textbf{Parent:} \texttt{0006}.
Node 0014 pivots to a market-scale-normalized level of news short interest, backfilled within subindustries, smoothed for 20 days, and held between valid updates; a small five-day return-reversal term times entry.
The positive short-interest sign is interpreted post hoc as crowded-short covering or a securities-lending premium, but the reviewer marks this explanation inadequate because the standard directional prior is bearish.
Subindustry neutralization, decay 20, and 2\% truncation produce Fitness 2.82, Sharpe 2.32, turnover 7.82\%, and drawdown 6.91\%.
The node also receives a regime warning because annual Fitness ranges from 0.95 to 7.08.

\paragraph{Node \texttt{0015}.}
\textbf{Artifact:} \url{0015-10-fast-persistent-short-acceleration};
\textbf{BRAIN ID:} \brainid{N1rJzP0L};
\textbf{Parent:} \texttt{0014}.
This child retains the persistent-short and five-day-reversal anchor, adds a small contrarian multifactor-acceleration sleeve, and changes portfolio controls to decay 8 and 0.75\% truncation.
The faster decay restores responsiveness, while the unusually tight cap limits the more concentrated composite.
The result is Fitness 3.24, Sharpe 2.61, turnover 11.94\%, and drawdown 6.59\%.
The report flags regime dependence: annual Fitness ranges from 0.91 to 6.87, so the in-sample improvement is not temporally uniform.

\paragraph{Node \texttt{0016}.}
\textbf{Artifact:} \url{0016-35-gaussian-macro-fund-final};
\textbf{BRAIN ID:} \brainid{XgnMr2Aa};
\textbf{Parent:} \texttt{0014}.
Node 0016 expands the persistent-short anchor with a capital-expenditure expectation surprise, a 720-day call-minus-put implied-volatility spread, and a 120-day call-implied-to-Parkinson-volatility ratio.
A Gaussian quantile map emphasizes cross-sectional tails, while trade-on-update logic prevents stale analyst data from generating artificial daily changes.
Subindustry neutralization, decay 36, and 0.455\% truncation yield Fitness 3.44, Sharpe 2.91, turnover 4.95\%, and drawdown 4.87\%.
The final truncation search is important to the passing Sharpe improvement, but annual Fitness still ranges from 0.97 to 7.04 and triggers a regime warning.

\paragraph{Node \texttt{0017}.}
\textbf{Artifact:} \url{0017-10-resid-all-snt-d25};
\textbf{BRAIN ID:} \brainid{xAkpRWlw};
\textbf{Parent:} \texttt{0006}.
The official name and report describe a relationship-return residual, but the evaluated \texttt{rettype=2} operator returns the 20-day regression slope of all-relationship returns on competitor returns.
The actual alpha therefore combines competitor-channel exposure, a social-sentiment z-score, and the systematic-risk term structure within relationship groups.
Subindustry neutralization, decay 25, and 8\% truncation produce raw Fitness 1.18, Sharpe 1.56, turnover 6.42\%, and drawdown 4.24\%.
Because the report's mechanism does not match the evaluated expression, the reviewer sets review Fitness to zero; annual Fitness from 0.05 to 2.46 also produces a regime warning.

\subsubsection{Branch rooted at 0011: event-conditioned close reversal}

\paragraph{Node \texttt{0011}.}
\textbf{Artifact:} \url{0011-8-raven-ratings-slower};
\textbf{BRAIN ID:} \brainid{rKl9GKXj};
\textbf{Parent:} \texttt{root}.
This root child uses the position of the close within the daily high--low range as a reversal signal and multiplies it by ranked absolute broad-news and analyst-ratings impact after 18-day backfill.
The absolute news values measure event importance without trusting the provider's direction, so the mechanism is a consequential-event filter on weak-close reversal.
With subindustry neutralization, decay 12, and 3\% truncation, the node records Fitness 1.02, Sharpe 2.21, turnover 54.81\%, and drawdown 3.10\%.
Its self-correlation exceeds the nominal threshold but passes through the more-than-10\% Sharpe-improvement exception.

\paragraph{Node \texttt{0012}.}
\textbf{Artifact:} \url{0012-4-annual80-reported60};
\textbf{BRAIN ID:} \brainid{np2m5EXa};
\textbf{Parent:} \texttt{0011}.
The child pivots from event reversal to two consensus-earnings-yield trends: a double-weight 80-day time-series rank of adjusted annual EPS over price and a 60-day rank of reported GAAP EPS over price.
The second sleeve checks whether the adjusted-EPS trend is supported by reported earnings rather than accounting exclusions alone.
Subindustry neutralization, decay 8, and 5\% truncation yield Fitness 1.69, Sharpe 1.98, turnover 12.23\%, and drawdown 4.35\%.
Annual Fitness ranges from 0.43 to 4.96, so the reviewer records regime dependence despite positive full-period metrics.

\paragraph{Node \texttt{0013}.}
\textbf{Artifact:} \url{0013-3-news-accel-direction-blend};
\textbf{BRAIN ID:} \brainid{2rL93WoJ};
\textbf{Parent:} \texttt{0011}.
Node 0013 preserves the parent's news-weighted weak-close reversal, increases its weight when multifactor acceleration has a large magnitude, and adds a contrarian rank of the 20-day mean acceleration direction.
The additive 0.25 floor prevents the event-reversal channel from disappearing when model activity is low.
With subindustry neutralization, decay 12, and 3\% truncation, the node reaches Fitness 1.31, Sharpe 2.52, turnover 43.40\%, and drawdown 3.68\%.
The result demonstrates that high Sharpe need not imply high Fitness when turnover remains materially above the denominator floor.

\subsubsection{Branch rooted at 0018: failed news-volume seed and successful option/fundamental descendants}

\paragraph{Node \texttt{0018}.}
\textbf{Artifact:} \url{0018-6-lowvol-group-zs};
\textbf{Simulation ID:} \brainid{qMlJ7V0O};
\textbf{Parent:} \texttt{root};
\textbf{Submission:} none.
This retained root child standardizes the negative 30-day decay of news-day volume surprise within sectors, buying quiet names and shorting unusually active ones.
The proposed mechanism is that quiet trading on news days signals lower information asymmetry and informed-trading risk.
Subindustry neutralization, decay 10, and 8\% truncation produce raw Fitness 0.45, Sharpe 1.40, and turnover 85.88\%.
The node is not submitted because it fails Fitness, maximum-turnover, concentration, and sub-universe checks; the reviewer also sets review Fitness to zero because its reported drawdown, margin, and long count do not match the final platform record.

\paragraph{Node \texttt{0019}.}
\textbf{Artifact:} \url{0019-12-tenor-blend-mean48};
\textbf{BRAIN ID:} \brainid{pwlL71Ex};
\textbf{Parent:} \texttt{0018}.
Node 0019 abandons news volume and sums put-minus-call implied-volatility disagreement at 30-, 60-, and 90-day tenors, averages it for 48 days, reverses the sign, and z-scores it within sectors.
Persistent demand for downside protection is treated as bearish, while agreement across tenors reduces dependence on one expiry.
Industry neutralization, decay 10, 8\% truncation, and NaN handling yield Fitness 4.73, Sharpe 3.03, turnover 5.08\%, and drawdown 11.15\%.
The 48-day window is explicitly tournament-selected and option-data coverage remains the main operational boundary, although all five in-sample years are positive.

\paragraph{Node \texttt{0021}.}
\textbf{Artifact:} \url{0021-28-forward-pcr-low-quartic};
\textbf{BRAIN ID:} \brainid{KPE0LnN1};
\textbf{Parent:} \texttt{0019}.
This child trades only when the 20-day mean of the 270-day put/call open-interest ratio is below one.
Within that call-dominant state, multi-tenor IV skew supplies direction, while industry-level skew magnitude and a 90/30 option-forward-curve deviation enter as squared and fourth-power confidence terms.
Industry neutralization, decay 12, and 8\% truncation produce the trace maximum: Fitness 9.50, Sharpe 3.48, turnover 7.93\%, and drawdown 19.50\%.
The threshold one has an economic interpretation, but the 48- and 60-day windows and the fitted powers create substantial selection risk; annual Fitness from 2.59 to 18.28 triggers a regime warning.

\paragraph{Node \texttt{0020}.}
\textbf{Artifact:} \url{0020-41-forward-1080-30-w60-d30};
\textbf{BRAIN ID:} \brainid{E5e5LxqP};
\textbf{Parent:} \texttt{0018}.
Node 0020 combines the sector-standardized difference between 1080- and 30-day synthetic option forwards with a contrarian composite-revision rank and activates a low-volatility tilt only in the smallest-capitalization 15\%.
The long-horizon forward curve is the principal information channel; the gated low-volatility term is designed to protect the higher-capitalization sub-universe check.
With no portfolio neutralization, decay 30, and 3\% truncation, the node records Fitness 2.24, Sharpe 1.69, turnover 2.01\%, and drawdown 13.66\%.
The reviewer flags the contrarian revision sign, the unexplained 0.35 centering constant, and regime dependence.

\paragraph{Node \texttt{0022}.}
\textbf{Artifact:} \url{0022-9-p8-s08};
\textbf{BRAIN ID:} \brainid{j2rExRxQ};
\textbf{Parent:} \texttt{0020}.
This child replaces the option-forward structure with annual and quarterly consensus-EPS-yield deviations from their 126-day means, plus a contrarian 60-day social-sentiment sleeve.
After z-scoring, an eighth signed power concentrates the book in the extreme tails and also reduces correlation with the earlier EPS-yield nodes.
Subindustry neutralization, decay 10, and 3\% truncation yield Fitness 2.77, Sharpe 2.23, turnover 8.76\%, and drawdown 7.89\%.
The power ladder is search-selected and the node remains regime-dependent, with annual Fitness falling to 0.13 in 2023 and reaching 5.81 in 2020.

\paragraph{Node \texttt{0023}.}
\textbf{Artifact:} \url{0023-10-quad-filter-dec20};
\textbf{BRAIN ID:} \brainid{gJ9Oz2Nv};
\textbf{Parent:} \texttt{0018}.
Node 0023 builds four separately standardized relationship-group sleeves: volatility-scaled weekly reversal, standardized year-over-year operating-income surprise, volatility-scaled competitor momentum, and gross-margin contraction.
Fine relationship groups are used for price sleeves and coarser groups for lower-coverage fundamentals; \texttt{filter=true} treats a missing sleeve as zero rather than removing the stock.
Subindustry neutralization, decay 20, and 8\% truncation produce Fitness 2.44, Sharpe 2.69, turnover 12.82\%, and drawdown 4.17\%.
The missing-value union is the largest documented improvement in this search, but the sub-universe Sharpe passes exactly at its limit and annual Fitness from 0.84 to 6.15 triggers a regime warning.

\paragraph{Node \texttt{0024}.}
\textbf{Artifact:} \url{0024-7-sept-cpiv-w100};
\textbf{BRAIN ID:} \brainid{9q70L2mr};
\textbf{Parent:} \texttt{0023}.
The final child retains all four parent sleeves and adds two option channels: a half-weight negative 500-day z-score of the 270-day put/call open-interest ratio and a full-weight 10-day decay of the 60-day call-minus-put implied-volatility spread.
Abnormally put-heavy positioning is bearish, while relatively expensive calls are interpreted as informed bullish demand that reaches the equity price with delay.
Subindustry neutralization, decay 20, and 8\% truncation yield Fitness 3.35, Sharpe 3.33, turnover 11.80\%, and drawdown 3.31\%.
Self-correlation with node 0023 is 0.8422 but passes through the Sharpe-improvement exception; the sub-universe Sharpe again passes exactly at its limit, and annual Fitness from 1.45 to 7.36 produces a regime warning.

\begin{table*}[t]
\centering
\scriptsize
\setlength{\tabcolsep}{4.0pt}
\renewcommand{\arraystretch}{1.08}
\begin{tabularx}{\textwidth}{L{0.055\textwidth}L{0.50\textwidth}Y}
\toprule
Node & Alpha construction & Economic interpretation \\
\midrule
\texttt{0001} &
\(R_s[-2Z_{20}(r)-M_3(r)/\sigma_{20}(r)]\) &
Volatility-scaled one- and three-day reversal: temporary liquidity pressure and overreaction should mean-revert within subindustries. \\
\cmidrule(lr){1-3}
\texttt{0002} &
\(R_s[Z_{252}(\widehat{\mathrm{EPS}}/P)]\) &
Forward earnings yield that is high relative to its own one-year history signals improving expectations relative to price. \\
\cmidrule(lr){1-3}
\texttt{0003} &
\(-R_s[M_{20}(\mathrm{analyst\ revision\ acceleration})]+R_s[-\Delta_5P/M_{20}(V)]\) &
Contrarian analyst-revision pressure is combined with a low-volume five-day price reversal. \\
\cmidrule(lr){1-3}
\texttt{0004} &
\(-R_s[M_{20}(\mathrm{multifactor\ acceleration})]+R_s[-\Delta_7P/M_{20}(V)]-1.3R_s[D_5(r^{cust})]+0.5R_s[-\sum_5 r]\) &
Slow multifactor and customer-return overreaction are reversed, with low-volume and conventional weekly price reversal as confirmation. \\
\cmidrule(lr){1-3}
\texttt{0005} &
\(-1.1R_s(\mathrm{static\ model}+\mathrm{acceleration})+R_s[(\mathrm{multi\mbox{-}horizon\ reversal})(0.5+R(V/\mathrm{ADV20}))]\) &
Contrarian model scores are paired with reversal whose confidence rises when volume is unusually high. \\
\cmidrule(lr){1-3}
\texttt{0006} &
\(R_G[D_{17}(r^{comp})]R[\log(1+n^{comp})]+0.11R_G(\mathrm{risk}_{360}-\mathrm{risk}_{90})\) &
Competitor-return information diffuses more reliably when supported by many links; a change in systematic risk supplies a small regime term. \\
\cmidrule(lr){1-3}
\texttt{0007} &
\(0.4\{R_s[M_5(\bar r_G)]-R_s[M_5(r)]\}+R_G[-\Delta_5P/M_{20}(V)]\) &
Buy product-cluster laggards when their peers have led, augmented by low-volume reversal within the relationship group. \\
\cmidrule(lr){1-3}
\texttt{0008} &
\(R_s[D_{20}(I^{news})]+0.32R_s[D_{24}(S^{M\&A})]+0.16R_i(\mathrm{risk}_{360}-\mathrm{risk}_{90})+0.18R_s[D_{20}(I^{ptg}_{C})]+0.07R_s[D_{20}(I^{ptg}_{N})]\) &
Persistent news impact, M\&A sentiment, price-target news, and a risk-regime shift are treated as complementary information-arrival channels. \\
\bottomrule
\end{tabularx}
\caption{Complete alpha catalog and economic interpretations, part 1 of 3.
All constructions are faithful operator-level compressions of the released FASTEXPR strings.}
\label{tab:alpha-catalog-a}
\end{table*}

\begin{table*}[t]
\centering
\scriptsize
\setlength{\tabcolsep}{4.0pt}
\renewcommand{\arraystretch}{1.08}
\begin{tabularx}{\textwidth}{L{0.055\textwidth}L{0.50\textwidth}Y}
\toprule
Node & Alpha construction & Economic interpretation \\
\midrule
\texttt{0009} &
\(\mathrm{0004}\) with an eight-day price leg, customer weight 1.2, and \(+1.5\,\mathbf{1}_{I^{business}\neq\mathrm{NaN}}R_s(I^{business})\) &
Business-news impact confirms the parent's reversal complex; the altered horizon and customer weight trade responsiveness against persistence. \\
\cmidrule(lr){1-3}
\texttt{0010} &
\(\mathrm{0004}+2.3R_s[\operatorname{backfill}_{180}(\mathrm{sales\ growth})]+0.8R_s[\operatorname{backfill}_{252}(\mathrm{book}/P)]\) &
Sales growth and book-to-price add fundamental growth and value to the parent's price, customer, and model reversals. \\
\cmidrule(lr){1-3}
\texttt{0011} &
\(R_s[C\,R(|\operatorname{backfill}_{18}I^{news}|)\,R(|\operatorname{backfill}_{18}I^{ratings}|)]\), \(C=(H+L-2P)/(H-L)\) &
Weak closes are interpreted as short-horizon liquidity pressure; broad and ratings-news magnitude selects reversals around consequential events. \\
\cmidrule(lr){1-3}
\texttt{0012} &
\(2R_s[\operatorname{tsrank}_{80}(\widehat{\mathrm{EPS}}_{adj}/P)]+R_s[\operatorname{tsrank}_{60}(\widehat{\mathrm{EPS}}_{GAAP}/P)]\) &
Adjusted and reported consensus earnings-yield trends jointly proxy improving fundamentals relative to price. \\
\cmidrule(lr){1-3}
\texttt{0013} &
\(R_s[CQR\{0.25+R(|A^{model}|)\}]-0.35R_s[M_{20}(A^{model})]\) &
News-conditioned weak-close reversal is emphasized when model acceleration is large, while its slow directional component is traded contrarily. \\
\cmidrule(lr){1-3}
\texttt{0014} &
\(R[H(\mathrm{observed},R\{M_{20}[\operatorname{groupfill}_{60}(SI/M_{252}\bar{SI}_{mkt})]\})+0.1R(-M_5r)]\) &
Persistent high short interest is treated as crowded-short covering or a securities-lending premium, plus entry-timing reversal; the reviewer flags this positive short-interest sign as under-justified. \\
\cmidrule(lr){1-3}
\texttt{0015} &
\(\mathrm{0014}-0.06R_s(A^{model})\), with faster portfolio decay and tighter truncation &
The persistent-short and reversal anchor is diversified by a contrarian multifactor-acceleration sleeve and stricter risk controls. \\
\cmidrule(lr){1-3}
\texttt{0016} &
\(\begin{aligned}
Q_NR[&R(\mathrm{0014})+0.08R_s(IV^{call}_{720}-IV^{put}_{720})\\
&+0.09R\{H(\Delta\widehat{CapEx}\neq0,\mathrm{CapEx\ surprise})\}\\
&+0.08R_s(IV^{call}_{120}/\sigma^{Park}_{120})]
\end{aligned}\) &
Short crowding and reversal are joined by capital-expenditure expectations, directional option demand, and an implied-versus-realized volatility premium; Gaussian mapping emphasizes cross-sectional tails. \\
\bottomrule
\end{tabularx}
\caption{Complete alpha catalog and economic interpretations, part 2 of 3.}
\label{tab:alpha-catalog-b}
\end{table*}

\begin{table*}[t]
\centering
\scriptsize
\setlength{\tabcolsep}{4.0pt}
\renewcommand{\arraystretch}{1.08}
\begin{tabularx}{\textwidth}{L{0.055\textwidth}L{0.50\textwidth}Y}
\toprule
Node & Alpha construction & Economic interpretation \\
\midrule
\texttt{0017}$^\dagger$ &
\(0.5R_G[\beta_{20}(r^{all\ rel},r^{comp})]+0.3R_G[Z_{20}(S^{social})]+0.2R_G(\mathrm{risk}_{360}-\mathrm{risk}_{90})\) &
The evaluated first term is exposure of all-relationship returns to competitor returns, not an orthogonal residual; sentiment and changing systematic risk provide confirmation. \\
\cmidrule(lr){1-3}
\texttt{0018}$^\dagger$ &
\(Z_{\mathrm{sector}}[-D_{30}(\texttt{news\_vol\_stddev})]\) &
Quiet news-day volume is interpreted as lower information asymmetry and informed-trading risk; the node was not submitted and its report contains mismatched secondary metrics. \\
\cmidrule(lr){1-3}
\texttt{0019} &
\(Z_{\mathrm{sector}}[-M_{48}\{\sum_{\tau\in\{30,60,90\}}(IV^{put}_{\tau}-IV^{call}_{\tau})\}]\) &
Persistent, multi-tenor demand for downside protection is bearish; low or reversed put--call disagreement is bullish relative to sector peers. \\
\cmidrule(lr){1-3}
\texttt{0020} &
\(R(F_{1080/30}+1.5C^{rev})-0.35+0.60\,\mathbf{1}_{R(cap)<0.15}V^{low}\) &
An upward long-horizon option-forward curve is combined with contrarian composite revisions and a small-cap-only low-volatility tilt; the revision sign and 0.35 centering constant are reviewer-flagged. \\
\cmidrule(lr){1-3}
\texttt{0021} &
\(H[M_{20}(PCR^{OI}_{270})<1,\ Z_{\mathrm{sector}}\{\operatorname{sp}_2(S)|I|^2|F|^4\}]\) &
During call-dominant long-tenor positioning, multi-tenor IV skew supplies direction while industry skew and the 90/30 forward curve supply nonlinear confidence; fitted powers are an explicit overfitting risk. \\
\cmidrule(lr){1-3}
\texttt{0022} &
\(\begin{aligned}
\operatorname{sp}_8 Z\{&
R[\Delta^{avg}_{126}(\widehat{EPS}_a/P)]\\
&+0.5R[\Delta^{avg}_{126}(\widehat{EPS}_q/P)]\\
&+0.8R[Z_{\mathrm{sector}}(-M_{60}S^{social})]\}
\end{aligned}\) &
Improving annual and quarterly consensus earnings yields are combined with depressed crowd sentiment, and the eighth power concentrates exposure in extreme composites. \\
\cmidrule(lr){1-3}
\texttt{0023} &
\(\begin{aligned}
&Z_{G_5}[-D_5(r)/\sigma_{60}(r)]\\
&+Z_{G_{20}}[\Delta_{252}OI/\sigma_{504}(\Delta_{252}OI)]\\
&+Z_{G_5}[D_5(r^{comp})/\sigma_{60}(r^{comp})]\\
&+Z_{G_{20}}[-\Delta_{252}GM]
\end{aligned}\) &
Within relationship peers, weekly reversal, standardized operating-income surprise, competitor momentum, and gross-margin mean reversion diversify four distinct information channels. \\
\cmidrule(lr){1-3}
\texttt{0024} &
\(\mathrm{0023}+0.5Z_{G_5}[-Z_{500}(PCR^{OI}_{270})]+Z_{G_5}[D_{10}(IV^{call}_{60}-IV^{put}_{60})]\) &
The four equity/fundamental sleeves gain two option channels: abnormally put-heavy positioning is bearish, while relatively expensive calls proxy informed bullish demand. \\
\bottomrule
\end{tabularx}
\caption{Complete alpha catalog and economic interpretations, part 3 of 3.
Daggers match the reviewer-invalidated reports in Table~\ref{tab:mcts-tree}.}
\label{tab:alpha-catalog-c}
\end{table*}

\FloatBarrier

\subsection{What this typical run demonstrates}

This trace shows how the program can move between genuinely different data families rather than merely retune one formula.
Some branches improve by adding information to a parent, as in the business-news and option overlays, while others abandon the parent's mechanism to escape a correlation or Fitness dead end.
The retained failure at node 0018 and the reviewer-invalidated explanation at node 0017 are part of the evidence: the tree records failed gates and mismatched reasoning rather than presenting only submitted winners.

At the same time, a typical sample run is not a repeated-run benchmark.
The high node-level submission rate, the best Fitness of 9.50, and all lineage-specific improvements describe this trace under one sequence of model states and platform responses.
The trace demonstrates that the complete workflow operates as specified under a fixed protocol, producing externally graded artifacts with full provenance.

\subsection{Behavior across calendar regimes}

Aggregate Fitness can conceal whether an alpha works steadily or earns most of its score in one market environment.
We therefore examine all nine nodes whose full-period Fitness exceeds 2.0.
For compactness, the five-number sequences below report annual Fitness in chronological order from 2019 through 2023.
The calendar years provide descriptive regime slices: roughly pre-pandemic, pandemic shock and rebound, reopening, monetary tightening, and the subsequent rebound.
They are not exogenous regime assignments.
These slices diagnose temporal concentration within the platform-graded window; they are not an author-defined substitute test set.

\paragraph{Persistent-short branch (0014--0016).}
The three related alphas have strikingly similar annual profiles.
Their annual Fitness sequences are
\[
\begin{aligned}
0014 &: (1.02,\,7.08,\,2.57,\,3.83,\,0.95),\\
0015 &: (0.95,\,6.87,\,3.56,\,4.25,\,0.91),\\
0016 &: (1.81,\,7.04,\,4.32,\,3.62,\,0.97).
\end{aligned}
\]
All three peak near Fitness 7 in 2020 and all fall to approximately 1 in 2023.
The acceleration and option/fundamental overlays improve some middle years, but they do not remove the shared temporal shape inherited from the persistent-short anchor.
Consequently, the rise in full-period Fitness from 2.82 at 0014 to 3.24 and 3.44 at its children should not be read as independent evidence of regime robustness.
The branch appears especially suited to the cross-sectional dislocations of 2020, while the positive high-short-interest sign remains both economically under-explained and empirically state dependent.

\paragraph{Option disagreement and forward-curve branch (0019--0021).}
Node 0019 is comparatively broad-based: its annual Fitness sequence is \((2.76,\,3.77,\,4.96,\,9.92,\,2.39)\), so every year exceeds 2 even though 2022 contributes about 41\% of total in-sample PnL.
Its descendant 0021 is stronger in every calendar year, at \((8.43,\,2.59,\,18.28,\,17.19,\,4.15)\), but the very large 2021--2022 values dominate its full-period Fitness of 9.50.
Thus, 0021 is not a one-year result, yet its nonlinear skew and forward-curve construction is clearly most effective in the 2021--2022 option regime.
The minimum annual Fitness of 2.59 is reassuring within the sample; the 7.1-fold gap between its best and worst years is not.

Node 0020 provides a useful contrast within the same broad option-data family.
Its annual Fitness sequence is
\[
(3.14,\,2.87,\,3.46,\,0.68,\,2.31).
\]
The long-horizon forward-curve, contrarian-revision, and small-cap low-volatility blend weakens precisely in 2022, when 0019 and 0021 are strongest.
This opposite 2022 behavior suggests that the program did not merely rediscover one generic option-market exposure.
It also shows why grouping alphas only by data source is insufficient: sign, tenor, conditioning, and the accompanying equity sleeves determine the regime profile.

\paragraph{Earnings-yield and sentiment tail signal (0022).}
Node 0022 records
\[
(1.69,\,5.81,\,2.95,\,5.01,\,0.13).
\]
It performs strongly in 2020 and 2022 but almost disappears in 2023.
The eighth signed power concentrates positions in extreme combinations of earnings-yield change and depressed sentiment; this improves full-period separation and lowers correlation with earlier earnings-yield alphas, but it also makes the result depend on years in which the composite produces sufficiently informative tails.
The 44.7-fold best-to-worst annual Fitness ratio is the most severe regime imbalance among the stronger nodes.

\paragraph{Relationship-peer branch (0023--0024).}
The four-sleeve parent 0023 records
\[
(1.45,\,0.84,\,6.15,\,1.41,\,3.54),
\]
with its largest payoff in 2021 and its weakest result in 2020.
Adding abnormal put/call positioning and call--put implied-volatility spread produces 0024's annual Fitness sequence:
\[
(1.93,\,1.45,\,7.36,\,3.22,\,3.46).
\]
The option overlay raises annual Fitness in 2019--2022 and is nearly neutral in 2023, where Fitness slips only from 3.54 to 3.46.
It also reduces the best-to-worst ratio from 7.3 to 5.1.
This is the clearest within-lineage evidence that a child improved temporal balance rather than merely increasing the best year, although 2021 remains dominant and the ratio still exceeds the reviewer's regime-warning threshold.

\paragraph{Cross-branch interpretation.}
The peaks are staggered rather than universal: the persistent-short branch is strongest in 2020, the relationship-peer branch in 2021, and the option-disagreement branch in 2022; node 0020 is unusually weak in that same 2022 environment.
This pattern is consistent with the tree discovering economically distinct exposures rather than variants of one hidden common score.
It is not, by itself, proof of portfolio diversification, because a combined book was not evaluated and all formulas were selected on the same five-year interval.
The main lesson is more modest: high aggregate Fitness should be read together with the annual path.
Nodes 0019 and 0024 show the most balanced positive annual evidence among the stronger alphas, whereas 0014--0016 and 0022 depend much more heavily on particular calendar regimes.

\paragraph{Constants and selection risk.}
Several formulas contain numerical constants that serve different purposes but create a common overfitting concern.
Some have an ex ante interpretation, such as the put/call threshold of one in 0021, which distinguishes call-dominant from put-dominant positioning, while others are portfolio weights or smoothing horizons chosen from tournament grids.
The unexplained centering constant \(0.35\) in 0020 is the clearest warning case: it mechanically changes the book's net exposure without an economic or calendar rationale.
The squared and fourth-power confidence terms in 0021, the eighth signed power in 0022, and the tenor, lookback, and sleeve weights in 0024 can likewise improve in-sample separation by concentrating on a small set of observations.
The program compared multiple nearby values before selecting these specifications; the selected constants are documented in the released reports so that every choice is inspectable.

\noindent\begin{minipage}{\columnwidth}
\subsection{Evidence index}

\centering
\footnotesize
\setlength{\tabcolsep}{3pt}
\begin{tabularx}{\columnwidth}{@{}L{0.42\columnwidth}L{0.20\columnwidth}Y@{}}
\toprule
Headline claim & PDF evidence & Release object \\
\midrule
Multiple users obtained SPECTACULAR & main \S5 & per-user appendix tables \\
Fitness 9.50, Sharpe 3.48 (\brainid{KPE0LnN1}) & main Table 3 & BRAIN record \\
Reviewer caught real defects (2 zero-score) & main \S5.3 & review blocks \\
10-worker deployment validated & main \S4.4 & scheduler event log \\
Two roles, 101 prompt lines & main \S4.2 & proposer/reviewer prompts \\
Complete prompt-to-factor trail & main \S5 & artifact manifest \\
\bottomrule
\end{tabularx}
\captionsetup{hypcap=false}
\captionof{table}{Evidence index: every headline result links to a specific, inspectable artifact bundle.}
\label{tab:evidence}
\end{minipage}

\subsection{Artifact completeness}

The accompanying release contains the proposer, reviewer, and dispatcher prompts; prompt history; scheduler code and complete MCTS state; all candidate reports; simulation inputs and responses; rankings; correlation records; submission checks; and final submission responses.
The node reports provide the literal executable strings underlying the compact constructions in Tables~\ref{tab:alpha-catalog-a}--\ref{tab:alpha-catalog-c}.
Together, these files connect each public result to the prompt that produced it, the candidates it beat, the reviewer text it received, and the platform record that supports its metrics.

The 30-paper audit provides a fixed comparison boundary \citep{yao2026beyond}.
No audited study is complete across its five reproducibility fields, even though 18 expose some artifacts.
AgonAlpha deliberately uses BRAIN's external grading rather than substituting an author-defined holdout from a local engine.
A chronological split tests temporal transfer conditional on the authors' backtest implementation.
BRAIN adds evaluator independence by controlling the data, simulator, metrics, gates, and grades.
Every submitted alpha then enters platform-run out-of-sample tracking.
On our side of that boundary, AgonAlpha releases the complete prompt-to-factor trail, exact production formulas, and returned platform records.
This is the first complete prompt-to-factor release in the audited LLM trading literature.

\FloatBarrier

\end{document}